\documentclass[11pt]{article}

\usepackage[utf8]{inputenc}
\usepackage[T1]{fontenc}
\usepackage[margin=1in]{geometry}
\usepackage{microtype}
\usepackage{amsmath,amssymb,amsfonts,amsthm,mathtools}
\usepackage{mathrsfs,bm,bbm,upgreek}
\usepackage{nicefrac,xspace}
\usepackage{graphicx}
\usepackage{booktabs,multirow,longtable,array,adjustbox}
\usepackage{float,wrapfig,subcaption,pdflscape}
\usepackage{algorithm,algorithmic}
\usepackage{enumerate}
\usepackage{xcolor}
\usepackage[authoryear,round]{natbib}
\usepackage[hidelinks]{hyperref}

\usepackage{amsmath,amsfonts,bm}

\def\eqref#1{equation~\ref{#1}}

\def\1{\bm{1}}

\DeclareMathAlphabet{\mathsfit}{\encodingdefault}{\sfdefault}{m}{sl}
\SetMathAlphabet{\mathsfit}{bold}{\encodingdefault}{\sfdefault}{bx}{n}

\title{Foundation Models for Event Prediction \\ Across Heterogeneous Event Logs}

\author{
\begin{tabular}{@{}cc@{}}
Fabian Spaeh & Sean Jaffe \\
Celonis & Celonis \\
\texttt{f.spaeh@celonis.com} & \texttt{s.jaffe@celonis.com} \\[0.75em]
Shandian Zhe & Bin Shen \\
Celonis & Celonis \\
\texttt{s.zhe@celonis.com} & \texttt{b.shen@celonis.com}
\end{tabular}
}
\date{October 2026}

\begin{document}

\maketitle

\begin{abstract}

Event logs are ubiquitous across real-world domains, including business processes, manufacturing workflows, healthcare, and many other operational systems. Yet event logs can differ substantially in their event semantics, label spaces, interaction patterns, and temporal scales, making it difficult to build a single foundation model that generalizes directly to previously unseen systems. We address this challenge with the \textbf{Foundation Event Predictor (FEP)}, a simple and scalable \textit{in-context} architecture that performs next-event prediction through learned, event-level nonparametric inference. Causal self-attention encodes context and target histories into key and query representations, respectively, while the observed next context events are encoded as values. Cross-attention then performs learned, relevance-weighted aggregation of next-event outcomes over all context transitions, allowing the entire nonparametric prediction procedure to be expressed as standard attention operations and efficiently parallelized with modern kernels such as FlashAttention. To promote cross-system generalization, we introduce permutation-robust embedding learning and random fixed embeddings to decouple event IDs from persistent semantics, together with empirical quantile transforms and robust normalization to handle diverse temporal scales. We pretrain FEP on 696 real-world event datasets spanning heterogeneous event systems and evaluate it on 50 held-out datasets unseen during pretraining. Across these datasets, FEP generalizes effectively to new event systems without parameter adaptation, frequently approaching the performance of strong models trained separately on each target dataset while substantially outperforming existing foundation-model baselines, particularly on challenging prediction tasks.
\end{abstract}

\section{Introduction}

Event logs arise across numerous real-world domains and modern operational systems. Examples include loan applications and claims processing in financial services, purchasing and replenishment in supply chains, order-to-cash and procure-to-pay workflows in enterprise systems, production and maintenance events in manufacturing, patient visits and clinical procedures in healthcare, and alarms and incidents in computing infrastructure. In each setting, an event log records sequences of discrete activities together with their occurrence times and potentially other descriptive information. Given an ongoing event sequence, predicting what event will occur next and when it will occur is a fundamental problem for understanding, monitoring, and anticipating the evolution of the underlying process. Accurate next-event prediction can support applications ranging from process monitoring, optimization, and resource planning to anomaly detection and proactive intervention~\citep{difrancescomarino2022predictive,ceravolo2024predictive,lee2022analysis}.

Building a general-purpose predictive model for event logs is challenging because event systems can differ substantially in their event semantics, label spaces, interaction patterns, sequence characteristics, and temporal scales. An event ID representing a purchase order in one system may have no relationship to the same ID in a clinical or manufacturing log. Consequently, most existing predictive approaches train a separate model for each event log, tying the learned predictor to the activity vocabulary, temporal scale, and dynamics of that particular system~\citep{mei2017neural,zuo2020transformer,Shchur2020IntensityFree,evermann2017predicting,tax2017predictive,bukhsh2021processtransformer}. Recent foundation-model approaches have begun to relax this restriction~\citep{berghaus2026fimpp,berti2026retrieval}, but generalizing directly across highly heterogeneous event systems remains challenging.

To address this problem, we propose \textbf{Foundation Event Predictor (FEP)}, a simple and scalable in-context architecture for next-event prediction across heterogeneous event logs. FEP treats observed context sequences as a collection of event-level predictive examples, each pairing an observed history with its subsequent event. Causal self-attention encodes context histories into keys and target histories into queries, while the observed subsequent context events form the corresponding values. Cross-attention then performs learned, relevance-weighted aggregation of next-event outcomes over the context, implementing an event-level nonparametric predictor. Its output is decoded into the next event type and inter-arrival time. In this way, FEP exploits the finest-grained supervision available in an event log---\textit{the multitude of observed history-to-next-event transitions}---and learns across systems how to transfer these predictive relationships to unseen systems. At the same time, expressing the entire nonparametric inference procedure through standard attention enables highly parallel computation over large context sets and direct use of optimized kernels such as FlashAttention~\citep{dao2022flashattention,dao2024flashattention2}.

To further improve generalization across systems with unrelated event vocabularies, we prevent FEP from associating fixed event IDs with persistent semantics. We introduce two complementary strategies: permutation training, which randomly permutes the mapping from event types to a shared learned embedding table in each training episode, and random fixed embeddings, which generate fresh non-trainable event-type representations in each episode. Both encourage FEP to infer predictive structure from the observed context rather than memorize dataset-specific label identities. To accommodate heterogeneous temporal scales and large within-log variation, we further introduce two context-dependent transformations of event inter-arrival times: an empirical quantile transform and a robust log-based normalization. FEP is trained using these normalized time representations to improve robustness across diverse temporal distributions.

We systematically evaluate FEP on a large and heterogeneous collection of real-world event logs, pretraining on 696 datasets and testing on 50 held-out datasets unseen during pretraining. We compare against strong event-prediction models trained separately on each held-out dataset as well as existing foundation models for event data, including FIM-PP~\citep{berghaus2026fimpp} and FM-v2~\citep{berti2026retrieval}. Across different difficulty levels, FEP consistently approaches the strongest local models in both event-type accuracy and inter-arrival-time prediction error while substantially outperforming competing foundation models. Moreover, FEP substantially outperforms an architecture-matched local counterpart on average, demonstrating the benefit of cross-dataset pretraining for generalization. We further conduct ablations on context size, the number of pretraining datasets, model dimension, and other design choices to characterize the factors driving FEP's performance, and analyze its inference efficiency and scalability.

\section{Related Work}

\label{sect:related}

\paragraph{Temporal point processes.}
Temporal point processes provide a classical framework for modeling irregular event sequences~\citep{daley2003introduction,hawkes1971spectra}. Hawkes processes capture self- and cross-excitation among event types, while modern neural point processes use recurrent or attention-based architectures to learn richer history-dependent dynamics, including RMTPP~\citep{du2016recurrent}, Neural Hawkes Processes~\citep{mei2017neural}, and Transformer-based models~\citep{zuo2020transformer,zhang2020self,yang2022transformer}. Intensity-free approaches instead directly parameterize next-event-time distributions~\citep{Shchur2020IntensityFree}. These models are typically trained separately for each event system and operate within a fixed event-type vocabulary, whereas we seek a single predictor that generalizes directly across heterogeneous, previously unseen event systems.

\paragraph{Predictive process monitoring.}
In process mining~\citep{vanderaalst2016process}, predictive process monitoring forecasts future behavior from prefixes of running cases. Deep models based on recurrent networks~\citep{evermann2017predicting,tax2017predictive} and Transformers~\citep{bukhsh2021processtransformer} have been developed for tasks such as next-activity, next-event-time, and remaining-time prediction. Despite strong predictive performance, these approaches generally train a separate model for each event log, tying the learned predictor to the activity vocabulary, temporal scale, and dynamics of that particular process.

\paragraph{Foundation inference for temporal point processes.}
The work most closely related to ours from the point-process perspective is FIM-PP~\citep{berghaus2026fimpp}, which pretrains a Foundation Inference Model on a large synthetic distribution of marked point processes and infers conditional intensity functions for unseen systems from context sequences. FIM-PP and FEP thus share the goal of transferring a single pretrained model to new event systems, but differ substantially in both training regime and inference mechanism. FIM-PP performs amortized inference of a Hawkes-like conditional intensity under a synthetic Hawkes-family prior. In contrast, FEP is trained across heterogeneous real-world event logs and treats each observed history-to-next-event transition as an in-context predictive example: context histories form attention keys and their observed next events form values. Cross-attention then learns to aggregate these event-level outcomes for direct next-event prediction, without explicitly estimating an intensity function.

\paragraph{Foundation models for predictive process monitoring.}
Berti and van der Aalst~\citep{berti2026incontext} study in-context prediction on unseen event logs without parameter updates.
Their model learns prefix embeddings and applies two separate nonparametric heads: prototype-based classification for next activity and kernel regression for remaining case time, i.e., the time until process completion rather than the time to the next activity. For each activity type, the prototype is computed as the mean embedding of support prefixes whose next activity has that type. The model uses balanced contexts or retrieval-based support selection. Their follow-up FM-v2~\citep{berti2026retrieval} makes retrieval central to both training and inference, learning representations explicitly optimized for nearest-neighbor selection of target-log support prefixes.

These works are related to FEP in their use of in-context examples and locally defined activity vocabularies. FEP, however, directly models the \emph{next event}, predicting both its event type and inter-arrival time, and unifies nonparametric inference through learned cross-attention over event-level history-to-next-event pairs. This formulation jointly learns the relevance function and outcome representation while exploiting the finest-grained supervision available in the log. Moreover, the nonparametric prediction  is expressed entirely through standard attention operations, enabling highly parallel batched computation using optimized attention implementations such as FlashAttention and FlashAttention-2~\citep{dao2022flashattention,dao2024flashattention2}.

\section{Methodology}
\label{sec:method}

\subsection{Problem Setting}

Suppose we are given $N$ event datasets
$\mathcal{P}=\{\mathcal{D}_1,\ldots,\mathcal{D}_N\}$ for pretraining, where each $\mathcal{D}_i$ ($1 \le i \le N$) contains event sequences from a particular system,
$$
\mathcal{D}_i=\{\Gamma_n^i\}_n,\qquad
\Gamma_n^i=(\mathbf e_{n1}^i,\mathbf e_{n2}^i,\ldots),
$$
and each event $\mathbf e_{nj}^i=(\Delta t_{nj}^i,s_{nj}^i)$ consists of an inter-arrival time and an event type. Different datasets may have unrelated event semantics and label spaces, different numbers of event types, distinct transition and interaction patterns, and drastically different temporal scales.

Our goal is to learn a single model that directly generalizes to previously unseen event systems without parameter adaptation. At inference time, a new system provides a \emph{context set} of exemplar sequences $\mathcal{D}^C=\{\Gamma_n^C\}_n$ and a set of ongoing query sequences $\mathcal{D}^{Q} = \{ \Gamma_n \}_n$. Given the observed history of a query sequence up to event $\mathbf e_m$, FEP uses $\mathcal{D}^C$ as in-context support to predict the next event
$\mathbf e_{m+1}=(\Delta t_{m+1},s_{m+1})$.

\subsection{Event-Level In-Context Prediction}

\begin{figure}
    \centering
    \includegraphics[trim={0 0.7cm 0 0.7cm},clip,width=0.95\linewidth]{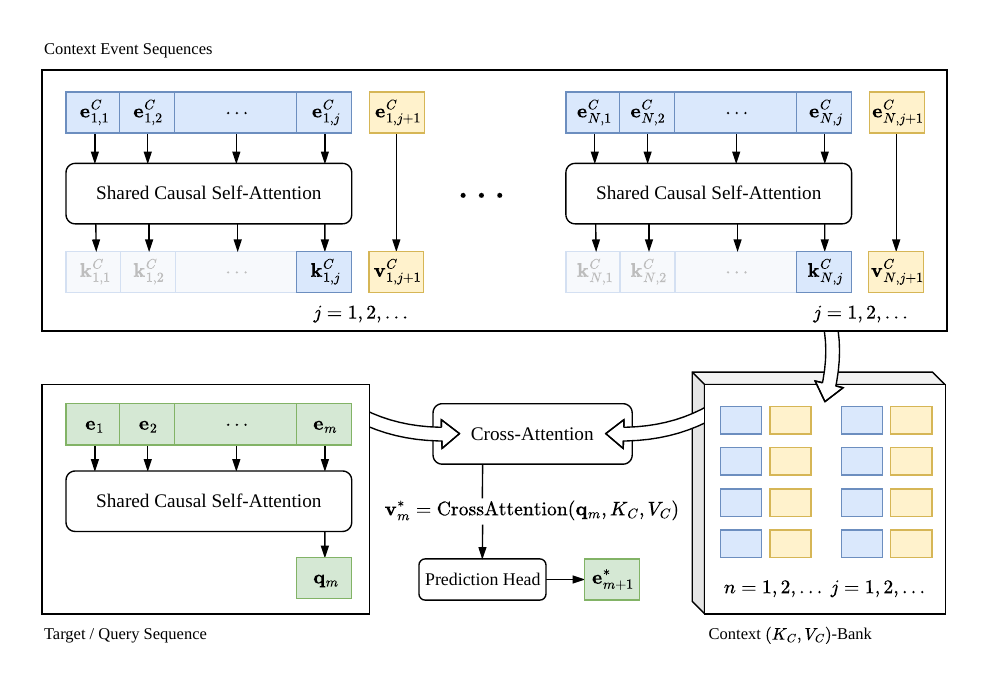}
    \caption{\small The architecture of Foundation Event Predictor (FEP). Context histories and observed next events form keys and values, respectively, while target histories form queries for cross-attention-based nonparametric prediction.}
    \label{fig:arch}

\end{figure}

The central idea of FEP is to decompose the context $\mathcal{D}^C$ into event-level predictive examples
$$
H_{nj}^C \longrightarrow \mathbf e_{n,j+1}^C,
$$
where $H_{nj}^C=(\mathbf e_{n1}^C,\ldots,\mathbf e_{nj}^C)$ denotes the history up to event $\mathbf e_{nj}^C$. FEP then performs learned nonparametric inference over this collection using standard attention. See Figure~\ref{fig:arch}.

Specifically, for each event $\mathbf e=(\Delta t,s)$ in $\mathcal{D}^C$ and $\mathcal{D}^Q$, we construct an input representation
$\mathbf h=[\alpha(\Delta t);A(s)]$, where $\alpha(\cdot)$ and $A(\cdot)$ denote time and event-type embeddings, respectively. The time embedding operates on the normalized time representation introduced later. We add sinusoidal positional encodings and encode both context and query sequences with a \textit{shared} causal Transformer. Specifically, letting
$\mathbf x_j=\mathbf h_j+\mathbf p_j$,
where $\mathbf p_j$ denotes the sinusoidal positional encoding at position $j$, we compute
$$
(\mathbf z_1,\ldots,\mathbf z_L)
=
\operatorname{CausalSelfAttention}(\mathbf x_1,\ldots,\mathbf x_L).
$$
The causal mask ensures that $\mathbf z_j$ summarizes only the history up to position $j$. We then construct
$$
\mathbf k_{nj}^C=W_K\mathbf z_{nj}^C,
\qquad
\mathbf q_m=W_Q\mathbf z_m,
$$
so that each context history is represented by a key and each target history by a query.
The corresponding value encodes the \emph{observed outcome} of the context history, i.e., the next event. Specifically, for
$\mathbf e_{n,j+1}^C=(\Delta t_{n,j+1}^C,s_{n,j+1}^C)$, we define
$$
\mathbf v_{nj}^C
=
[\mathbf v_{\Delta,nj}^C;\mathbf v_{s,nj}^C],
$$
where
$$
\mathbf v_{\Delta,nj}^C
=
\operatorname{MLP}_{V,\Delta}\!\left(
\alpha(\Delta t_{n,j+1}^C)\right),
\qquad
\mathbf v_{s,nj}^C
=
\operatorname{MLP}_{V,s}\!\left(
\operatorname{CenterOneHot}(s_{n,j+1}^C)\right).
$$
Here, $\operatorname{CenterOneHot}(\cdot)$ denotes the centered one-hot encoding of the event type. Thus, $(\mathbf k_{nj}^C,\mathbf v_{nj}^C)$ explicitly represents a history-to-next-event example.

Collecting all context keys and values into $K_C$ and $V_C$, FEP predicts the next event of the target history through cross-attention,
$$
\mathbf v_m^*
=
\operatorname{CrossAttention}(\mathbf q_m,K_C,V_C).
$$
For a single attention head, this corresponds to
$$
\mathbf v_m^*
=
\sum_{n,j} a_{m,nj}\mathbf v_{nj}^C,
\qquad
a_{m,nj}
=
\frac{
\exp(\mathbf q_m^\top\mathbf k_{nj}^C/\sqrt d)}
{\sum_{n',j'}
\exp(\mathbf q_m^\top\mathbf k_{n'j'}^C/\sqrt d)},
$$
where $d$ is the key/query dimension. Hence, attention performs a learned, relevance-weighted nonparametric aggregation over all observed context transitions: more relevant context histories receive larger weights, whereas irrelevant transitions can be effectively ignored.

Writing $\mathbf v_m^*=[\mathbf z_\Delta^*;\mathbf z_s^*]$, two lightweight prediction heads decode
$$
\widehat y_{m+1}
=
\operatorname{MLP}_{\Delta}(\mathbf z_\Delta^*),
\qquad
\boldsymbol{\ell}_{m+1}
=
\operatorname{MLP}_{s}(\mathbf z_s^*),
$$
where $\widehat y_{m+1}$ is the predicted normalized inter-arrival time and $\boldsymbol{\ell}_{m+1}$ contains logits over the target system's event types. We optimize
$$
\mathcal L
=
\operatorname{CE}(\boldsymbol{\ell}_{m+1},s_{m+1})
+
\lambda\cdot
\operatorname{MSE}(\widehat y_{m+1},y_{m+1}),
$$
where $y$ denotes the normalized inter-arrival-time representation defined below.

This architecture directly exploits the finest-grained predictive information available in the context set---the multitude of observed history-to-next-event transitions. Importantly, both causal history encoding and context aggregation are standard attention operations: keys and queries for many sequences can be computed in parallel, while cross-attention over large context sets can exploit efficient implementations such as FlashAttention~\citep{dao2022flashattention,dao2024flashattention2}.

\subsection{Robust Representation of Time}

Time scales can differ dramatically across event systems and even within a single process. For example, activities in a loan application may be separated by seconds or minutes, while later approval stages may take days. We therefore normalize inter-arrival times using statistics computed exclusively from the context set $\mathcal D^C$.

We consider two alternatives. The first constructs an empirical quantile transform $F_C$ from context inter-arrival times and maps $y=F_C(\Delta t)\in[0,1]$.
FEP predicts $y$, and the prediction is mapped back to the original time scale using $F_C^{-1}$.

The second performs robust normalization in log time. For each context inter-arrival time $\Delta t_i$, let  $r_i=\log(\Delta t_i+\epsilon)$, and define: $m_C=\operatorname{median}_i(r_i),\,
\operatorname{MAD}_C=
\operatorname{median}_i\left|r_i-m_C\right|$.
We then set
$$
s_C=\max\!\left(\tau\,\operatorname{MAD}_C,s_{\min}\right),
\qquad
y=
\frac{\log(\Delta t+\epsilon)-m_C}{s_C},
$$
where $\tau = 1/\Phi^{-1}(0.75) \approx 1.4826$ makes the scaled MAD a consistent estimator of the standard deviation under a Gaussian distribution~\citep{rousseeuw1993alternatives}. This normalization remains robust to extreme waiting times. Predictions are mapped back to the original time scale through
$\widehat{\Delta t}=\exp\!\left(m_C+s_C\widehat y\right)-\epsilon$.

\subsection{Removing Spurious Semantics from Event IDs}

A shared event-ID embedding table is problematic across heterogeneous logs because the same numerical ID can represent completely unrelated activities in different systems. We therefore prevent FEP from relying on persistent ID semantics and instead encourage it to infer event relationships from the observed context.

We study two strategies. \textbf{Permutation training} maintains a shared learned embedding table while randomly permuting the correspondence between event types and embedding IDs in each training episode. The embedding table thus provides a shared representational basis without assigning persistent semantics to particular IDs. Alternatively, \textbf{random fixed embeddings} generate a fresh set of random event-type vectors for each episode. These vectors distinguish event types within the current system but are neither optimized nor preserved across episodes, thereby eliminating persistent semantics associated with particular event IDs.

\subsection{Training Across Event Systems}
Training proceeds episodically across the pretraining datasets.
For each event system, a context loader samples sequences used to construct $(K_C,V_C)$, while an inference loader samples query sequences and their next-event targets. FEP computes query representations, performs cross-attention against the context using FlashAttention, evaluates the prediction loss, and immediately backpropagates the loss for the current query batch. The corresponding computation graph is then released before moving to the next system, reducing peak memory consumption. Gradients are accumulated across systems, followed by a parameter update at the end of each episode. We further parallelize training across GPUs using PyTorch DistributedDataParallel (DDP)~\citep{li2020pytorchdistributed}. The training procedure is summarized in Algorithm~\ref{alg:fep} in the appendix.

\section{Experiments}
\label{sec:experiments}

\subsection{Generalization Performance}

To evaluate generalization to unseen event systems, we test FEP on 50 public real-world datasets spanning diverse domains, including business processes, finance, healthcare, mobility, environmental monitoring, and public administration. Table~\ref{tab:test-domain-statistics} in the appendix summarizes the datasets by domain, while Table~\ref{tab:test-only-dataset-catalog} provides the full list and descriptions. FEP is pretrained on 696 other real-world datasets collected from public repositories such as \href{https://data.4tu.nl/}{4TU.ResearchData} and \href{https://www.kaggle.com/datasets}{Kaggle}. We ensure that the 50 test datasets do not overlap with the pretraining data in dataset or process identity, including duplicate or derived versions of the same underlying event log. A summary of the pretraining datasets is provided in Table~\ref{tab:pretrain-domain-statistics}, with further details in Appendix~\ref{sect:dataset-info}.

\paragraph{Competing Methods.}
We compare against strong task-specific local methods for event prediction and existing foundation models for event data. The local baselines include a Transformer designed specifically for next-event prediction, GRU~\citep{cho-etal-2014-learning}, Neural Hawkes Process~\citep{mei2017neural}, Transformer Hawkes Process~\citep{zuo2020transformer}, and Intensity-Free Temporal Point Process (IFTPP)~\citep{Shchur2020IntensityFree}. Each method is trained separately on the training set of each dataset. For each model, we select the best hyperparameter configuration and training epoch based on validation performance and report the corresponding test performance.

We also compare FEP with existing foundation models for event data, including FM-v2~\citep{berti2026retrieval} and FIM-PP~\citep{berghaus2026fimpp}. For these models, we provide in-context examples from the training sets and evaluate predictive performance on the corresponding test sets. We use the published checkpoints for \href{https://drive.google.com/file/d/1S34YJXsP0ZhRFaqicdlKwp0CckMvM-1K/view}{FM-v2} and \href{https://huggingface.co/FIM4Science/fim-pp/tree/9b43be14fc922bb389c873bf242fa9dd642662a5}{FIM-PP}. Finally, to quantify the benefit of cross-dataset pretraining, we evaluate an architecture-matched local counterpart of FEP, denoted as FEP-Local. Full implementation, validation, and configuration details are provided in Appendix~\ref{apx:exp}.

\paragraph{Metrics.}
For each test dataset, we report next-event-type prediction accuracy and normalized mean absolute error (NMAE) for next-event inter-arrival time. We define $\mathrm{NMAE}=
\frac{\mathrm{MAE}}
{\mu_{\Delta}+1\text{ second}}$,
where $\mu_{\Delta}$ is the mean ground-truth inter-arrival time on the test set and MAE is measured in seconds. The one-second constant is included for numerical robustness.

\paragraph{Evaluation.}
For each dataset, we use the newest 15\% of traces, ordered by sequence start time, as the test set. We randomly select 15\% of the remaining traces for validation and use the rest for training.

To characterize performance across different levels of empirical prediction difficulty, we group the 50 datasets into \textit{easy}, \textit{moderate}, and \textit{hard} categories based on the test performance of GRU and Neural Hawkes, whose hyperparameters and training epochs are selected independently on the validation sets. We choose these two models because they are well established and represent distinct modeling paradigms: GRU performs direct sequence prediction, whereas Neural Hawkes models event dynamics through a temporal point process. For each dataset, we average the two models' event-type accuracies and, separately, their NMAEs. Datasets are ranked in descending order of accuracy and ascending order of NMAE, and the two ranks are averaged to obtain an overall difficulty score. We then divide the datasets into three approximately equal-sized groups according to this score. Note that this grouping is used only for stratified analysis; all 50 datasets are included in the overall evaluation.

Since average metrics can be disproportionately influenced by a small number of catastrophic predictions, we additionally report Bradley--Terry scores for both event-type accuracy and NMAE~\citep{bradley1952rank}. For each dataset, methods are compared pairwise, with higher accuracy (or lower NMAE) counted as a win. We aggregate the resulting pairwise outcomes across datasets and jointly estimate a latent performance parameter for each method under the Bradley--Terry model. A higher score therefore indicates that a method wins pairwise comparisons more consistently across datasets, rather than being dominated by a small number of unusually large gains or failures.

\begin{table*}[t]
\centering
\footnotesize
\caption{\small Generalization performance by dataset difficulty, averaged over five runs across 50 test datasets. The best results among local methods and foundation models are highlighted in bold. FEP (Quantile) uses permutation training with the empirical quantile transform, while FEP (Robust log) uses random fixed embeddings with robust log-time normalization; all four combinations are reported in Appendix~\ref{sect:more-res}. FEP-Local is the architecture-matched local counterpart of FEP (Quantile), trained separately on each target dataset. FIM-PP supports at most 22 activities; we therefore retain the 21 most frequent activities and merge the rest into an ``other'' activity. FIM-PP (FT) denotes FIM-PP further fine-tuned on the training set of each target dataset. FM-v2 is omitted from NMAE because it predicts remaining case time, while Transformer Hawkes and IFTPP are omitted because their NMAEs exceed $10{,}000$ across all difficulty levels.}
\label{tab:difficulty-results}

\begin{subtable}{\textwidth}
\centering
\caption{Next-event-type accuracy.}
\label{tab:difficulty-accuracy}
\begin{tabular}{lccc}
\toprule
Method & Easy & Moderate & Hard \\
\midrule
Transformer & $\pmb{0.83 \pm 0.00}$ & $\pmb{0.66 \pm 0.00}$ & $\pmb{0.38 \pm 0.03}$ \\
GRU & $\pmb{0.83 \pm 0.00}$ & $\pmb{0.66 \pm 0.01}$ & $0.37 \pm 0.01$ \\
Neural Hawkes & $\pmb{0.83 \pm 0.00}$ & $0.65 \pm 0.00$ & $0.31 \pm 0.01$ \\
Transformer Hawkes & $0.76 \pm 0.02$ & $0.57 \pm 0.05$ & $0.24 \pm 0.01$ \\
IFTPP & $0.81 \pm 0.01$ & $0.62 \pm 0.01$ & $0.31 \pm 0.03$ \\
FEP-Local & $0.77 \pm 0.02$ & $0.60 \pm 0.02$ & $0.26 \pm 0.04$ \\
\midrule
FM-v2 & $0.77 \pm 0.00$ & $0.55 \pm 0.00$ & $0.24 \pm 0.00$ \\
FIM-PP & $0.30 \pm 0.01$ & $0.38 \pm 0.00$ & $0.14 \pm 0.00$ \\
FIM-PP (FT) & $0.57 \pm 0.04$ & $0.53 \pm 0.02$ & $0.32 \pm 0.05$ \\
FEP (Quantile) & $\pmb{0.82 \pm 0.00}$ & $\pmb{0.64 \pm 0.00}$ & $\pmb{0.35 \pm 0.00}$ \\
FEP (Robust log) & $\pmb{0.82 \pm 0.00}$ & $\pmb{0.64 \pm 0.00}$ & $\pmb{0.35 \pm 0.00}$ \\
\bottomrule
\end{tabular}
\end{subtable}

\bigskip

\begin{subtable}{\textwidth}
\centering
\caption{Next-event inter-arrival-time NMAE.}
\label{tab:difficulty-nmae}
\begin{tabular}{lccc}
\toprule
Method & Easy & Moderate & Hard \\
\midrule
Transformer & $0.52 \pm 0.36$ & $\pmb{0.54 \pm 0.01}$ & $11.22 \pm 17.78$ \\
GRU & $\pmb{0.28 \pm 0.00}$ & $\pmb{0.54 \pm 0.01}$ & $\pmb{\phantom{0}0.83 \pm \phantom{0}0.01}$ \\
Neural Hawkes & $0.38 \pm 0.03$ & $0.76 \pm 0.04$ & $\phantom{0}1.36 \pm \phantom{0}0.81$ \\
FEP (Local) & $0.31 \pm 0.04$ & $1.75 \pm 0.87$ & $\phantom{0}0.88 \pm \phantom{0}0.05$ \\
\midrule
FIM-PP & $0.79 \pm 0.00$ & $0.83 \pm 0.00$ & $\phantom{0}1.27 \pm \phantom{0}0.00$ \\
FIM-PP (FT) & $0.45 \pm 0.02$ & $0.82 \pm 0.00$ & $\phantom{0}1.07 \pm \phantom{0}0.01$ \\
FEP (Quantile) & $\pmb{0.30 \pm 0.00}$ & $\pmb{0.53 \pm 0.00}$ & $\pmb{\phantom{0}0.83 \pm \phantom{0}0.00}$ \\
FEP (Robust log) & $\pmb{0.30 \pm 0.00}$ & $0.55 \pm 0.00$ & $\phantom{0}0.84 \pm \phantom{0}0.00$ \\
\bottomrule
\end{tabular}
\end{subtable}
\end{table*}

\paragraph{Results and Discussion.}
Tables~\ref{tab:difficulty-accuracy} and~\ref{tab:difficulty-nmae} report the average event-type accuracy and inter-arrival-time NMAE across the three difficulty categories. The four combinations of event-ID representation and time normalization perform similarly; we therefore report two representative variants here and provide all four configurations in Appendix~\ref{sect:more-res}.
\begin{table}[t]
\setlength{\tabcolsep}{4pt}
\centering
\footnotesize
\caption{\small Overall pairwise ratings across 50 test datasets, reported as Bradley--Terry scores (higher is better). Scores are computed jointly from pairwise comparisons among all local, foundation, and fine-tuned models.}
\label{tab:benchmark-BT-score}
\begin{tabular}{lrr}
\toprule
Method & MAE score & Accuracy score  \\
\midrule
Transformer & $1644$ & $\pmb{1664}$ \\
GRU & $\pmb{1690}$ & $1631$ \\
Neural Hawkes & $1518$ & $1619$ \\
Transformer Hawkes & $1377$ & $1422$ \\
IFTPP & $1169$ & $1540$ \\
FEP (Local) & $1641$ & $1436$ \\
\midrule
FM-v2 & \textemdash & $1382$ \\
FIM-PP & $1396$ & $1176$ \\
FIM-PP (FT) & $1433$ & $1332$ \\
FEP (Quantile) & $\pmb{1675}$ & $\pmb{1561}$ \\
FEP (Robust log) & $1631$ & $1539$ \\

\bottomrule
\end{tabular}
\end{table}
FEP consistently achieves the best performance among the foundation models in both event-type accuracy and time prediction. In particular, FIM-PP remains substantially behind FEP even after target-specific fine-tuning. Compared with task-specific local models, FEP is only slightly behind, and in several cases on par with, the strongest local baselines---Transformer for event-type accuracy and GRU for NMAE. In contrast, FEP-Local shows substantially larger gaps from the best local models across all three difficulty categories. Cross-dataset pretraining markedly improves over FEP-Local and brings FEP close to the strongest task-specific models, demonstrating the benefit of learning predictive structure across heterogeneous event systems.

\begin{figure}
\centering
\begin{tabular}{cc}
    \includegraphics[width=0.40\linewidth]{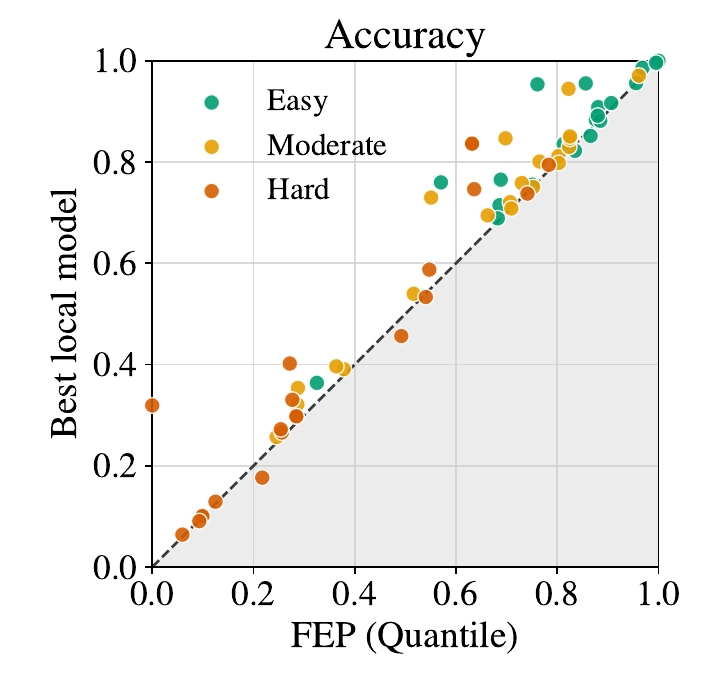}
     &
     \includegraphics[width=0.40\linewidth]{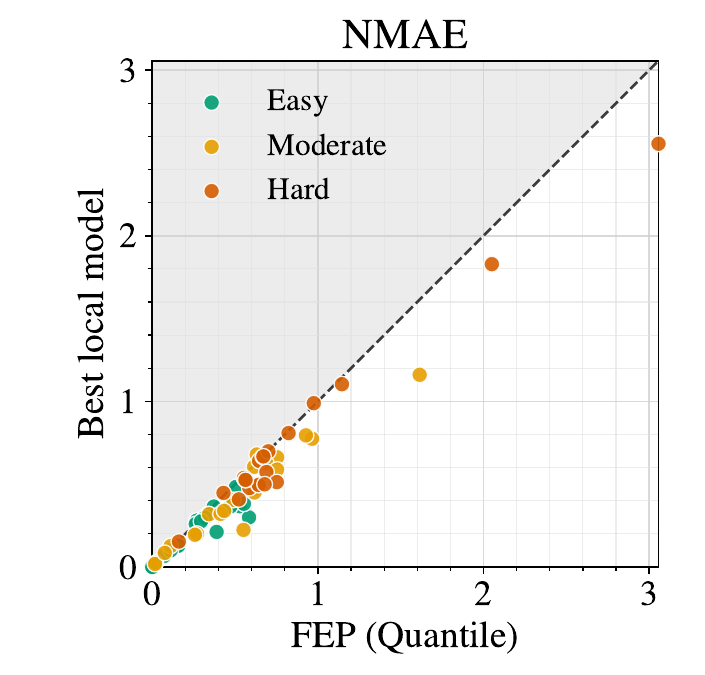}
\end{tabular}
    \caption{\small Dataset-wise comparison between FEP and the best locally trained model. Colors indicate empirical dataset difficulty, and the dashed line denotes equal performance. The shaded region indicates the side of the diagonal where FEP performs better.}
    \label{fig:fep-vs-best-local}
\end{figure}

We next examine performance on individual datasets. Figure~\ref{fig:fep-vs-best-local} compares FEP with the best local model separately for each test dataset, where the identity of the best local model may \textit{vary across datasets}. Most points lie close to the diagonal for both accuracy and NMAE, showing that FEP generally approaches the performance of the best locally trained model despite requiring no target-specific parameter adaptation. A small number of datasets exhibit larger gaps.

Finally, the pairwise comparisons summarized by the Bradley--Terry scores in Table~\ref{tab:benchmark-BT-score} show that FEP achieves the highest scores among foundation models and remains competitive with the local baselines. Results on six representative datasets are further provided in Appendix~\ref{sect:more-res} (Table~\ref{tab:selected-dataset-results}), where FEP matches or outperforms local models in either event-type accuracy or NMAE and outperforms the competing foundation models in all but one case, where FM-v2 attains the same event-type prediction accuracy.

\subsection{Ablation Studies}

We conduct a series of ablation studies to examine the factors affecting FEP's performance. For computational efficiency, unless otherwise stated, we pretrain on 33 datasets randomly sampled from the full pretraining pool and report accuracy and NMAE averaged over a fixed subset of 23 held-out datasets from the 50 test datasets.

\begin{figure}[t]
    \centering
    \begin{tabular}{cc}
        \begin{subfigure}[t]{0.48\linewidth}
            \centering
            \includegraphics[width=\linewidth]{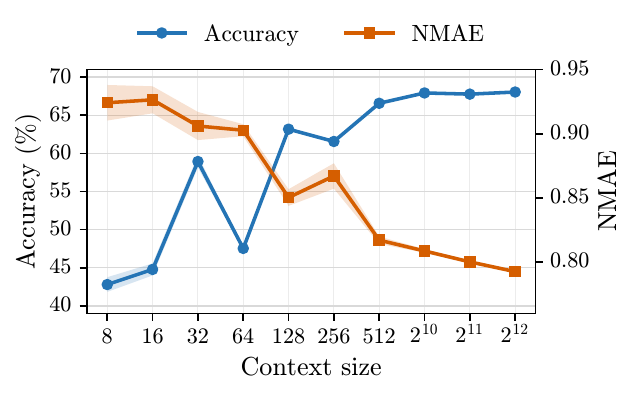}
            \caption{\small Context size \textit{vs.} performance.}
            \label{fig:ablation-context-size}
        \end{subfigure}
        &
        \begin{subfigure}[t]{0.48\linewidth}
            \centering
            \includegraphics[width=\linewidth]{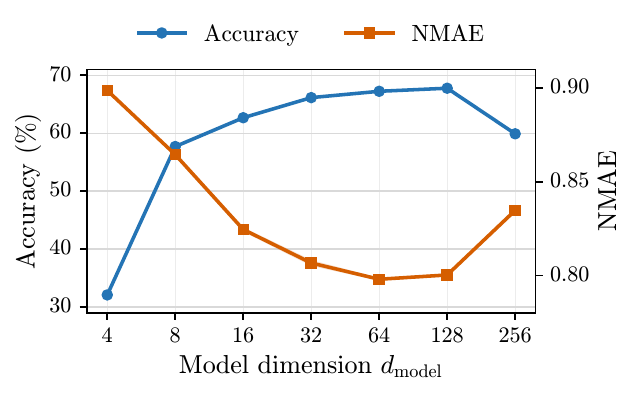}
            \caption{Model dimension \textit{vs.} performance.}
            \label{fig:ablation-model-dimension}
        \end{subfigure}
    \end{tabular}
    \caption{\small Ablation results for context size and model dimension.}
    \label{fig:ablation-context-model}
\end{figure}

\paragraph{Context size.}
Figure~\ref{fig:ablation-context-size} shows the effect of varying the number of context traces used during both training and inference. Increasing the context size typically improves both accuracy and NMAE,  with diminishing returns as more context is provided. Because each trace contains multiple events, even a moderate number of traces yields a large event-level context for cross-attention. The continued improvement with increasing context size suggests that FEP can effectively exploit additional context without evident attention dilution over the range considered.

\paragraph{Number of Pretraining Datasets.}
Table~\ref{tab:ablation-num-training-datasets} examines the effect of pretraining scale. With only 3 or 10 datasets, FEP generalizes poorly to unseen event systems, including unstable time prediction when trained on 10 datasets. Increasing the pretraining set to 30 datasets substantially improves both event-type accuracy and NMAE. Performance changes only modestly from 30 to 50 datasets, while pretraining on the full collection of 696 datasets provides a further improvement. These results indicate that exposure to diverse event systems is important for cross-system generalization, with gains becoming smaller as the number of pretraining datasets increases.

\paragraph{Model Size.}
Figure~\ref{fig:ablation-model-dimension} and Table~\ref{tab:ablation-depth} examine the effects of model dimension and depth. FEP performs well with relatively shallow networks: increasing depth from one to a few layers provides modest gains, whereas excessive depth degrades event-type accuracy. Increasing the model dimension yields substantial improvements up to moderate widths, with performance largely saturating around $d_{\mathrm{model}}=64$--$128$; increasing it further to 256 degrades both accuracy and NMAE. Overall, these results suggest that FEP does \textit{not} require an excessively large backbone to achieve strong cross-system generalization.

Additional ablation results, including the full combinations of event-ID representation and time normalization, as well as an analysis of inference efficiency and scalability, are provided in Appendix~\ref{sect:more-res}.

\begin{table}[t]
  \caption{\small Ablation results for the number of pretraining datasets and model depth. $\infty$ denotes NMAE larger than $10{,}000$.}
  \centering
  \footnotesize
  \label{tab:ablation-pretrain-depth}
	\vspace{-0.4cm}
  \begin{subtable}[t]{0.48\linewidth}
      \centering
      \caption{}
      \label{tab:ablation-num-training-datasets}
      \begin{tabular}{lcc}
        \toprule
        \#Pretraining datasets & Accuracy (\%) & NMAE \\
        \midrule
        3   & $39.09 \pm 0.00$ & $0.95 \pm 0.00$ \\
        10  & $49.52 \pm 0.05$ & $\infty$ \\
        30  & $67.45 \pm 0.01$ & $0.80 \pm 0.00$ \\
        50  & $67.59 \pm 0.02$ & $0.80 \pm 0.00$ \\
        696 & $\pmb{68.70 \pm 0.02}$ & $\pmb{0.77 \pm 0.00}$ \\
        \bottomrule
      \end{tabular}
  \end{subtable}\hspace{0.06\linewidth}\begin{subtable}[t]{0.36\linewidth}
      \centering
      \caption{}
      \label{tab:ablation-depth}
      \begin{tabular}{lcc}
        \toprule
        \#Layers & Accuracy (\%) & NMAE \\
        \midrule
        1 & $66.77 \pm 0.03$ & $0.83 \pm 0.00$ \\
        2 & $\pmb{67.91 \pm 0.02}$ & $0.80 \pm 0.00$ \\
        4 & $67.77 \pm 0.03$ & $0.80 \pm 0.00$ \\
        6 & $67.70 \pm 0.04$ & $\pmb{0.79 \pm 0.00}$ \\
        8 & $62.34 \pm 0.04$ & $0.81 \pm 0.00$ \\
        \bottomrule
      \end{tabular}
  \end{subtable}
\end{table}

\section{Limitations}

A primary limitation of the current FEP framework is that it only models event types and timestamps, without incorporating additional event attributes or sequence-level metadata. In practice, event logs often contain rich auxiliary information that can provide strong predictive signals and further improve next-event prediction. However, such information is itself highly heterogeneous across datasets: feature schemas, value ranges, categorical cardinalities, availability, and noise levels can differ substantially from one event log to another. Designing a representation that can flexibly incorporate these heterogeneous attributes while preserving cross-system generalization therefore remains a challenging open problem. We leave this extension to future work.

\section{Conclusion}
We have presented FEP, an in-context foundation model that generalizes to unseen heterogeneous event logs. The model is pretrained and evaluated on diverse real-world event data. Its next-event prediction performance without parameter adaptation is encouraging, often approaching strong local models trained specifically on target datasets. Future work will incorporate heterogeneous event features, expand the pretraining pool with synthetic data, and investigate uncertainty calibration for both prediction and process simulation.

\newpage
\bibliographystyle{plainnat}
\bibliography{reference_updated}

\newpage
\appendix

\section{Experimental Details}\label{apx:exp}

\paragraph{FEP configuration and training.}
The shared causal self-attention and cross-attention modules described in the main text use a hidden dimension of 128, four attention heads, four attention layers, a feed-forward dimension of 512, and dropout of 0.05. The maximum training pool for an individual event system is 2,560 traces; for larger systems, a new subset is sampled in each training episode. The context and query batch sizes are 512 and 128 traces, respectively. On each worker, four context minibatches are combined to form a context set of up to 2,048 traces for cross-attention.

We optimize FEP using AdamW~\citep{loshchilov2018decoupled} with a learning rate of $10^{-3}$ and weight decay of $10^{-5}$. Matrix multiplications use TF32 where supported, while the forward pass uses bfloat16 automatic mixed precision. We train FEP for 500 episodes and use the resulting pretrained model for all evaluations reported in the paper.

Training is distributed across four NVIDIA RTX PRO 6000 Blackwell Server Edition GPUs on a Linux machine with 96 Intel Xeon Scalable CPU cores and 1024 GiB of main memory. Each worker independently constructs its own context set of up to 2,048 traces, processes its own query batch, and computes local gradients. Gradients are accumulated across event systems and then averaged across workers by PyTorch DistributedDataParallel (DDP)~\citep{li2020pytorchdistributed} before the parameter update.
Cross-attention over each worker's context bank is implemented using FlashAttention-2.

\paragraph{Task-specific baselines.}
We train the local Transformer, GRU, Neural Hawkes, Transformer Hawkes, and IFTPP independently on the training split of each target dataset. For the Transformer and GRU, each activity is represented by a learned embedding. To accommodate multi-scale and potentially multimodal temporal distributions, the Transformer decomposes raw inter-arrival times into multiple time units, including days, hours, and minutes, following recent work on process-event prediction~\citep{spaeh2026universalmultimodaltraceformerintegrating}. The GRU instead applies a logarithmic transformation to raw inter-arrival times. Both models encode the resulting temporal representations using Time2Vec~\citep{kazemi2019} followed by an MLP, and combine them with the activity embeddings to form event tokens.
Both models use a linear classification head for next-activity prediction. The Transformer uses a factorized probabilistic time head over the day, hour, and minute components, whereas the GRU predicts a log-normal distribution for the next inter-arrival time.

All local models use AdamW, a batch size of 256, weight decay of 0.03, and gradient clipping with maximum norm 1.0. Training is conducted on the training split of each test dataset for at most 1,000 epochs, with early stopping after 15 consecutive validation evaluations without improvement. We evaluate six hyperparameter configurations for each model. Across the search grids, hidden dimensions are 64 or 128, learning rates are $10^{-3}$ or $3\cdot10^{-4}$, and dropout is 0 or 0.1 where applicable. The Transformer uses two or four layers with feed-forward dimensions of 128 or 256; GRU and Neural Hawkes use one or two layers; Transformer Hawkes uses two or four layers; and IFTPP uses one or two layers with three or five mixture components. Neural Hawkes, Transformer Hawkes, and IFTPP rescale time using the 90th percentile of the training inter-arrival times.

For each model--dataset pair, validation loss controls early stopping. Among the retained epochs, we rank validation activity accuracy in descending order and validation time MAE in ascending order, and select the epoch with the best equal-weight rank sum. We apply the same rank-sum criterion to select the hyperparameter configuration. The test set is evaluated only after model selection is complete. Final metrics are computed from five independently trained instances of the selected configuration, using different random initializations.

The Transformer and GRU are implemented directly in PyTorch. For Neural Hawkes, Transformer Hawkes, and IFTPP, we use the corresponding implementations from EasyTPP~\citep{xueeasytpp}.

\paragraph{Foundation-model baselines.}
FM-v2 and the zero-shot FIM-PP baseline are evaluated without updating their pretrained parameters.\footnote{
We use the publicly available checkpoints for
\href{https://drive.google.com/file/d/1S34YJXsP0ZhRFaqicdlKwp0CckMvM-1K/view}{FM-v2} and
\href{https://huggingface.co/FIM4Science/fim-pp/tree/9b43be14fc922bb389c873bf242fa9dd642662a5}{FIM-PP}.
}
FM-v2 performs retrieval over prefixes from the target training partition, using the 20 nearest support prefixes for each query. It receives at most 2,048 support traces. For our evaluation, we use its next-activity predictions; its temporal prediction targets remaining case time rather than next-event inter-arrival time.

FIM-PP receives the same maximum number of support traces and predicts both the next activity and its time. The fine-tuned FIM-PP variant follows the model's target-system adaptation procedure using at most 500 traces sampled from the target training partition; the validation and test partitions remain unchanged. We use the official fine-tuning implementation released together with the FIM-PP checkpoint.

The released FIM-PP implementation supports at most 22 activity classes. For datasets containing more classes, we retain the 21 most frequent activities in the sampled support set and map all remaining activities to a single ``other'' class. We therefore report FIM-PP accuracy in this 22-class label space. Because collapsing multiple original activities into one class makes the prediction problem easier, this preprocessing favors FIM-PP in the comparison.

\paragraph{In-context evaluation.}
We evaluate on 50 test datasets, whose aggregate statistics are reported in Table~\ref{tab:test-domain-statistics}. FEP and the frozen foundation-model baselines use at most 2,048 traces from the target training partition as in-context support. When the training partition contains more than 2,048 traces, we evaluate five independently sampled context subsets; otherwise, each method uses all available training traces once. The fine-tuned FIM-PP evaluation uses five independently sampled fine-tuning sets subject to its 500-trace limit. Thus, variation for task-specific local models reflects independent training runs, whereas variation for frozen foundation models reflects sampling of the target-system context.

\begin{algorithm}
\caption{Episodic pretraining of FEP}
\label{alg:fep}
\begin{algorithmic}[1]
\FOR{each training episode}
\STATE Initialize accumulated gradients to zero
\FOR{each event dataset $\mathcal D_m\in\mathcal P$}
\STATE Sample context sequences $\mathcal D^C$ and query batch $\mathcal B_{\rm Q}$
\STATE Causally encode $\mathcal D^C$ to construct context keys $K_C$ and next-event values $V_C$
\STATE Causally encode $\mathcal B_{\rm Q}$ to obtain query representations $Q$
\STATE Compute $\operatorname{CrossAttention}(Q,K_C,V_C)$ and next-event predictions
\STATE Compute $\mathcal L$ and backpropagate to accumulate gradients
\STATE Release the current computation graph
\ENDFOR
\STATE Update model parameters using the accumulated gradients
\ENDFOR
\end{algorithmic}
\end{algorithm}

\newpage
\section{Additional Results}
\label{sect:more-res}
\subsection{Inference Efficiency and Scalability}
Table~\ref{tab:ablation-inference-scaling} evaluates the efficiency and scalability of FEP's cached-context nonparametric inference. Because support-conditioned inference is expressed entirely as cross-attention, the same exact computation can be implemented either with standard attention, which explicitly materializes the attention score and probability matrices, or with FlashAttention-2, which performs exact attention using a fused, memory-efficient kernel. For FlashAttention-2, we explicitly select PyTorch's \texttt{SDPBackend.FLASH\_ATTENTION} backend.

\begin{table}
  \centering
  \footnotesize

  \caption{Inference efficiency and scalability of standard attention and FlashAttention-2 (FA2) for cached-context nonparametric inference.
  Each context sequence and query trace has length 100, yielding 99 history-to-next-event transitions or predictions.
  Peak memory includes model parameters, cached context representations,
  query execution, and output predictions.}
  \label{tab:ablation-inference-scaling}
  \resizebox{\linewidth}{!}{\begin{tabular}{llllllll}
    \toprule
    & & \multicolumn{2}{c}{Latency (ms)} &
    \multicolumn{2}{c}{Queries/s} & \multicolumn{2}{c}{Peak memory (GiB)} \\
    \cmidrule(lr){3-4}\cmidrule(lr){5-6}\cmidrule(lr){7-8}
    \# Context seq. & \# Context K/V tokens & Standard & FA2 & Standard & FA2 & Standard & FA2 \\
    \midrule
    8      & 792       & 2.118  & 2.182  & 472.2 & 458.2 & 0.039 & 0.038 \\
    16     & 1,584     & 2.124  & 2.185  & 470.7 & 457.6 & 0.042 & 0.040 \\
    32     & 3,168     & 2.120  & 2.191  & 471.7 & 456.5 & 0.050 & 0.045 \\
    64     & 6,336     & 2.112  & 2.188  & 473.5 & 457.0 & 0.064 & 0.055 \\
    128    & 12,672    & 2.121  & 2.196  & 471.5 & 455.5 & 0.094 & 0.073 \\
    256    & 25,344    & 2.115  & 2.189  & 472.7 & 456.7 & 0.153 & 0.104 \\
    512    & 50,688    & 2.131  & 2.214  & 469.2 & 451.7 & 0.267 & 0.169 \\
    1,024  & 101,376   & 2.209  & 2.187  & 452.7 & 457.3 & 0.497 & 0.297 \\
    \midrule
    2,048  & 202,752   & 3.228  & 2.606  & 309.8 & 383.7 & 0.962 & 0.557 \\
    4,096  & 405,504   & 5.336  & 3.852  & 187.4 & 259.6 & 1.885 & 1.078 \\
    8,192  & 811,008   & 9.469  & 6.187  & 105.6 & 161.6 & 3.729 & 2.114 \\
    12,288 & 1,216,512 & 13.548 & 8.470  & 73.8  & 118.1 & 5.576 & 3.156 \\
    16,384 & 1,622,016 & 17.616 & 10.733 & 56.8  & 93.2  & 7.420 & 4.193 \\
    \bottomrule
  \end{tabular}}
\end{table}

\begin{table}
\centering
\footnotesize
\caption{Relative improvement of FlashAttention-2 over standard attention at large context sizes. Positive values indicate improvement.}
\label{tab:fa2-relative-gains}
\begin{tabular}{rccc}
\toprule
\# Context seq. (tokens) &
Latency reduction &
Throughput increase &
Memory reduction \\
\midrule
2,048 (203K)   & 19.3\% & 23.9\% & 42.1\% \\
4,096 (406K)   & 27.8\% & 38.5\% & 42.8\% \\
8,192 (811K)   & 34.7\% & 53.0\% & 43.3\% \\
12,288 (1.2M)  & 37.5\% & 60.0\% & 43.4\% \\
16,384 (1.6M)  & 39.1\% & 64.1\% & 43.5\% \\
\bottomrule
\end{tabular}
\end{table}

Measurements use a batch size of one, BF16 autocast, PyTorch 2.14.0 with CUDA 13.0, and one NVIDIA RTX PRO 6000 Blackwell Server Edition. Each query is a length-100 trace producing 99 next-event predictions. We perform seven independent timing runs, each averaging 50 CUDA-event-timed queries after 20 warm-up queries, and report the median across runs. Fixed-length synthetic inputs control trace length while preserving production tensor shapes and execution kernels. Context encoding is excluded because the encoded context is cached and reused across queries.

At small context sizes, both implementations have nearly constant latency because fixed execution costs dominate, with standard attention approximately 3--4\% faster through 512 context sequences. FlashAttention-2 nevertheless already provides substantial memory savings in this regime. As the context grows, the latency trend reverses: FlashAttention-2 becomes faster at approximately 1,024 sequences, and its advantage increases steadily thereafter. At 2,048 sequences, FlashAttention-2 reduces latency by 19.3\% and increases throughput by 23.9\%; at 8,192 sequences, these gains grow to 34.7\% and 53.0\%, respectively. At the largest tested context of 16,384 sequences (1.62M context tokens), FlashAttention-2 reduces latency from 17.616 to 10.733 ms, increases throughput from 56.8 to 93.2 queries/s, and reduces peak memory from 7.420 to 4.193 GiB---corresponding to improvements of 39.1\%, 64.1\%, and 43.5\%, respectively (Table~\ref{tab:fa2-relative-gains}).

These results directly support the scalability motivation of FEP: expressing support-conditioned nonparametric inference as standard attention allows FEP to directly benefit from optimized attention implementations. At fixed query length, exact cross-attention still requires computation linear in the context size, but FlashAttention-2 avoids materializing the full attention matrices, yielding increasingly substantial runtime and memory savings as the context grows.

\begin{table}
  \centering
  \footnotesize

  \caption{Effect of event-ID representation under the ablation protocol.}
  \label{tab:ablation-permutation}
  \begin{tabular}{lcc}
    \toprule
    Setting & Accuracy (\%) & NMAE \\
    \midrule
    Random fixed        & $67.77 \pm 0.03$ & $0.80 \pm 0.00$ \\
    Learned + permutation    & $\pmb{67.95 \pm 0.03}$ & $\pmb{0.79 \pm 0.00}$ \\
    Learned, no permutation & $64.14 \pm 0.03$ & $0.83 \pm 0.00$ \\
    \bottomrule
  \end{tabular}
  \medskip
\end{table}
\subsection{Ablations on Event-ID Representations}

Following the ablation protocol described in the main paper, all variants in Table~\ref{tab:ablation-permutation} are pretrained on 33 datasets and evaluated on the 23 held-out datasets, with inter-arrival times represented using robust log normalization.
We compare three strategies for constructing the event-type embeddings used by the causal self-attention layers. The first uses random fixed embeddings, freshly sampled in each training episode and not optimized. The second uses learned embeddings with permutation training, where the correspondence between event types and embedding IDs is randomly permuted in each episode. The third uses learned embeddings without permutation.

Both random fixed embeddings and learned embeddings with permutation training achieve strong and similar performance, with permutation training performing slightly better overall. In contrast, learned embeddings without permutation reduce accuracy by nearly four percentage points and also increase NMAE. These results indicate that preventing persistent semantics from being associated with numerical event IDs is important for cross-system generalization, while both random fixed embeddings and permutation training provide effective solutions.

\begin{table*}
\centering
\small

\caption{
Performance of all four FEP configurations across dataset difficulty levels. The configurations combine two inter-arrival-time transformations (quantile and robust log) with two event-ID representation strategies (permutation training and random fixed embeddings).
}
\label{tab:four-comb-performance}

\begin{subtable}{\textwidth}
\centering
\caption{Next-event-type accuracy.}
\label{tab:four-comb-accuracy}
\begin{tabular}{ll|ccc}
\toprule
$\Delta t$ Transform & Type Embedding & Easy & Moderate & Hard \\
\midrule
Quantile & Permutation training & $0.82 \pm 0.00$ & $0.64 \pm 0.00$ & $ {0.35 \pm 0.00}$ \\
Robust log & Permutation training & $0.81 \pm 0.00$ & $0.64 \pm 0.00$ & $0.34 \pm 0.00$ \\
Quantile & Random fixed & $0.82 \pm 0.00$ & $0.64 \pm 0.00$ & $0.35 \pm 0.00$ \\
Robust log & Random fixed & $ {0.82 \pm 0.00}$ & $0.64 \pm 0.00$ & $0.35 \pm 0.00$ \\
\bottomrule
\end{tabular}
\end{subtable}

\bigskip

\begin{subtable}{\textwidth}
\centering
\caption{Next-event inter-arrival-time NMAE.}
\label{tab:four-comb-nmae}
\begin{tabular}{ll|ccc}
\toprule
$\Delta t$ Transform & Type Embedding & Easy & Moderate & Hard \\
\midrule
Quantile & Permutation training & $0.30 \pm 0.00$ & ${0.53 \pm 0.00}$ & $ {\phantom{0}0.83 \pm \phantom{0}0.00}$ \\
Robust log & Permutation training & ${0.29 \pm 0.00}$ & $0.54 \pm 0.00$ & $\phantom{0}0.84 \pm \phantom{0}0.00$ \\
Quantile & Random fixed & $0.31 \pm 0.00$ & $0.54 \pm 0.00$ & $\phantom{0}0.85 \pm \phantom{0}0.00$ \\
Robust log & Random fixed & $0.30 \pm 0.00$ & $0.55 \pm 0.00$ & $\phantom{0}0.84 \pm \phantom{0}0.00$ \\
\bottomrule
\end{tabular}
\end{subtable}

\end{table*}

\subsection{Combinations of Event-ID Representation and Time Normalization}

Table~\ref{tab:four-comb-performance} reports all four combinations of event-ID representation and time normalization, with FEP pretrained on the full set of 696 datasets. The four configurations achieve close performance in both event-type accuracy and inter-arrival-time NMAE across all difficulty levels, with no clear winner. This suggests that FEP is robust to the particular choice between the two proposed event-ID representations and time-normalization strategies.

\subsection{Dataset-wise Effect of Cross-Dataset Pretraining}
\begin{figure}
    \centering
    \includegraphics[width=0.4\linewidth]{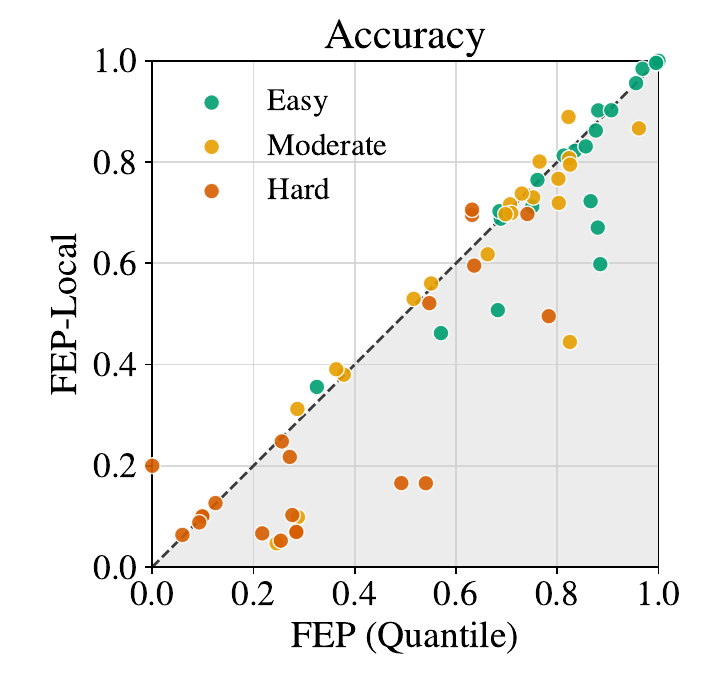}~
    \includegraphics[width=0.4\linewidth]{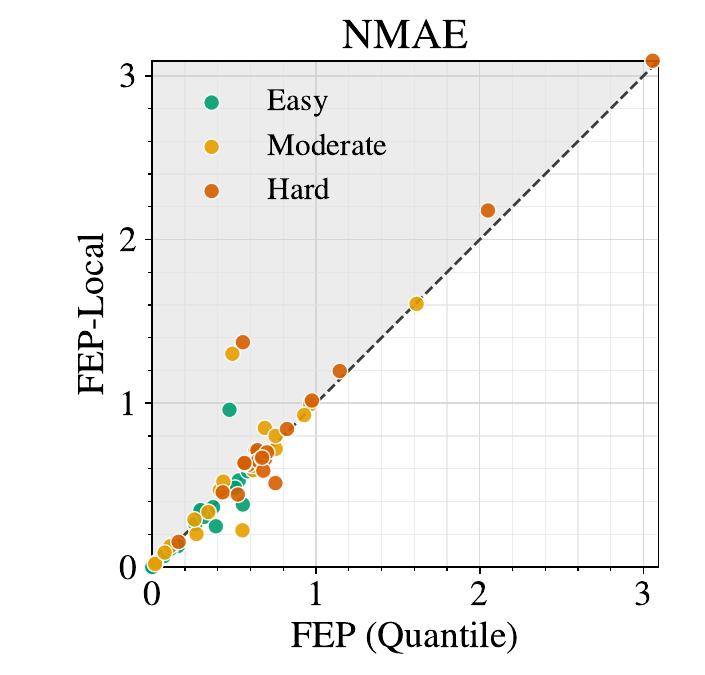}\\
    \caption{Dataset-wise comparison of FEP (Quantile) and FEP-Local on the 50 held-out datasets. Colors denote dataset difficulty, and the dashed line indicates equal performance. Points below the diagonal favor FEP for accuracy, whereas points above the diagonal favor FEP for NMAE. The FEP-favoring region is shaded in both panels. }
    \label{fig:dataset-comp-with-fep-local}
\end{figure}

Figure~\ref{fig:dataset-comp-with-fep-local} compares FEP with FEP-Local on each test dataset. For event-type accuracy, most points lie below the diagonal, indicating higher accuracy for FEP, with substantial improvements on many datasets. For NMAE, most points lie above or close to the diagonal, likewise indicating lower or comparable time-prediction error for FEP. These dataset-wise results show that the gains from cross-dataset pretraining are broadly distributed rather than driven by only a few datasets. Consistent with this observation, Table~\ref{tab:difficulty-results} in the main paper shows that FEP substantially outperforms FEP-Local on average across all three difficulty levels. Together, these results demonstrate the benefit of cross-dataset pretraining for generalization to unseen event systems.

\begin{table*}
\setlength{\tabcolsep}{5pt}
\centering
\footnotesize
\caption{
Accuracy and NMAE on six representative datasets, with two selected from each difficulty level. Results are reported analogously to Table~\ref{tab:difficulty-results}; NMAE values above $10{,}000$ are denoted by $\infty$.
}
\label{tab:selected-dataset-results}
\begin{tabular}{lcccccc}
\toprule
& \multicolumn{2}{c}{Medical Appointments} & \multicolumn{2}{c}{Beijing Air Quality} & \multicolumn{2}{c}{NASA SMAP/MSL} \\
\cmidrule(lr){2-3}\cmidrule(lr){4-5}\cmidrule(lr){6-7}
Method & Accuracy & NMAE & Accuracy & NMAE & Accuracy & NMAE \\
\midrule
Transformer & $\pmb{0.96 \pm 0.00}$ & $0.30 \pm 0.00$ & $0.76 \pm 0.02$ & $\pmb{0.68 \pm 0.01}$ & $0.76 \pm 0.01$ & $1.04 \pm 0.01$ \\
GRU & $\pmb{0.96 \pm 0.00}$ & $0.30 \pm 0.01$ & $\pmb{0.80 \pm 0.02}$ & $\pmb{0.68 \pm 0.01}$ & $\pmb{0.79 \pm 0.00}$ & $\pmb{0.99 \pm 0.00}$ \\
Neural Hawkes & $\pmb{0.96 \pm 0.00}$ & $0.31 \pm 0.01$ & $0.78 \pm 0.01$ & $0.72 \pm 0.02$ & $0.79 \pm 0.00$ & $1.04 \pm 0.05$ \\
Trans. Hawkes & $\pmb{0.96 \pm 0.00}$ & $0.33 \pm 0.03$ & $0.74 \pm 0.02$ & $0.77 \pm 0.06$ & $0.26 \pm 0.06$ & $1.32 \pm 0.19$ \\
IFTPP & $\pmb{0.96 \pm 0.00}$ & $\infty$ & $0.76 \pm 0.01$ & $0.79 \pm 0.02$ & $0.78 \pm 0.01$ & $1.51 \pm 0.55$ \\
FEP-Local & $\pmb{0.96 \pm 0.00}$ & $\pmb{0.28 \pm 0.00}$ & $0.72 \pm 0.02$ & $0.71 \pm 0.01$ & $0.50 \pm 0.22$ & $1.02 \pm 0.01$ \\
\midrule
FM-v2 & $\pmb{0.96 \pm 0.00}$ & \textemdash & $0.36 \pm 0.00$ & \textemdash & $0.40 \pm 0.00$ & \textemdash \\
FIM-PP & $0.01 \pm 0.00$ & $0.36 \pm 0.00$ & $0.30 \pm 0.00$ & $0.88 \pm 0.00$ & $0.39 \pm 0.00$ & $2.19 \pm 0.00$ \\
FIM-PP (FT) & $0.47 \pm 0.00$ & $0.34 \pm 0.00$ & $0.43 \pm 0.00$ & $0.91 \pm 0.00$ & $0.75 \pm 0.01$ & $1.52 \pm 0.04$ \\
FEP (Quantile) & $\pmb{0.96 \pm 0.00}$ & $\pmb{0.27 \pm 0.00}$ & $\pmb{0.80 \pm 0.00}$ & $0.63 \pm 0.00$ & $\pmb{0.78 \pm 0.00}$ & $\pmb{0.98 \pm 0.00}$ \\
FEP (Robust log) & $\pmb{0.96 \pm 0.00}$ & $0.28 \pm 0.00$ & $0.77 \pm 0.00$ & $\pmb{0.61 \pm 0.00}$ & $0.77 \pm 0.00$ & $1.02 \pm 0.00$ \\

\bottomrule
\end{tabular}

\medskip

\begin{tabular}{lcccccc}
\toprule
& \multicolumn{2}{c}{USGS Water} & \multicolumn{2}{c}{India Air Quality} & \multicolumn{2}{c}{US Drought Monitor} \\
\cmidrule(lr){2-3}\cmidrule(lr){4-5}\cmidrule(lr){6-7}
Method & Accuracy & NMAE & Accuracy & NMAE & Accuracy & NMAE \\
\midrule
Transformer & $0.81 \pm 0.01$ & $0.02 \pm 0.00$ & $\pmb{0.72 \pm 0.00}$ & $0.09 \pm 0.12$ & $\pmb{0.74 \pm 0.00}$ & $0.82 \pm 0.01$ \\
GRU & $\pmb{0.82 \pm 0.00}$ & $\pmb{0.00 \pm 0.00}$ & $\pmb{0.72 \pm 0.00}$ & $\pmb{0.03 \pm 0.00}$ & $0.73 \pm 0.01$ & $\pmb{0.81 \pm 0.01}$ \\
Neural Hawkes & $\pmb{0.82 \pm 0.00}$ & $0.50 \pm 0.19$ & $\pmb{0.72 \pm 0.00}$ & $0.83 \pm 0.08$ & $0.73 \pm 0.01$ & $0.88 \pm 0.05$ \\
Trans. Hawkes & $\pmb{0.82 \pm 0.01}$ & $0.09 \pm 0.05$ & $\pmb{0.72 \pm 0.00}$ & $0.30 \pm 0.21$ & $0.73 \pm 0.01$ & $0.82 \pm 0.10$ \\
IFTPP & $0.70 \pm 0.11$ & $0.01 \pm 0.00$ & $0.71 \pm 0.01$ & $0.04 \pm 0.01$ & $\pmb{0.74 \pm 0.00}$ & $0.88 \pm 0.01$ \\
FEP-Local & $\pmb{0.82 \pm 0.01}$ & $\pmb{0.00 \pm 0.00}$ & $\pmb{0.72 \pm 0.00}$ & $\pmb{0.03 \pm 0.00}$ & $0.70 \pm 0.01$ & $0.84 \pm 0.01$ \\
\midrule
FM-v2 & $0.47 \pm 0.00$ & \textemdash & $0.58 \pm 0.00$ & \textemdash & $0.37 \pm 0.00$ & \textemdash \\
FIM-PP & $0.12 \pm 0.00$ & $0.99 \pm 0.00$ & $0.59 \pm 0.01$ & $0.66 \pm 0.00$ & $0.37 \pm 0.00$ & $1.09 \pm 0.00$ \\
FIM-PP (FT) & $0.49 \pm 0.00$ & $0.99 \pm 0.00$ & $0.69 \pm 0.02$ & $0.62 \pm 0.03$ & $0.32 \pm 0.02$ & $1.14 \pm 0.00$ \\
FEP (Quantile) & $\pmb{0.84 \pm 0.00}$ & $\pmb{0.00 \pm 0.00}$ & $\pmb{0.71 \pm 0.00}$ & $\pmb{0.02 \pm 0.00}$ & $\pmb{0.74 \pm 0.00}$ & $\pmb{0.82 \pm 0.00}$ \\
FEP (Robust log) & $\pmb{0.84 \pm 0.00}$ & $0.01 \pm 0.00$ & $\pmb{0.71 \pm 0.00}$ & $0.03 \pm 0.00$ & $0.71 \pm 0.00$ & $\pmb{0.82 \pm 0.00}$ \\

\bottomrule
\end{tabular}
\end{table*}

\begin{table*}
\centering
\footnotesize
\caption{Summary of the real-world datasets used for pretraining and evaluation, grouped by domain.
Trace length and duration denote the mean number of events and total duration per case, respectively; all values are unweighted averages of per-dataset statistics within each domain.
\textbf{Environmental monitoring is entirely held out from the pretraining pool}, providing evaluation on a previously unseen domain.}
\label{tab:dataset-domain-statistics}

\begin{subtable}{\textwidth}
\centering
\caption{Statistics of the 696 pretraining datasets.}
\label{tab:pretrain-domain-statistics}
\setlength{\tabcolsep}{4pt}
\begin{tabular}{lrrrrr}
\toprule
Domain & \# & \#Cases & \#Activities & Trace length & Duration (days) \\
\midrule
Commerce, finance, and organizations
& 102 & 74\,036.17 & 20.74 & 172.70 & 529.61 \\
Healthcare and life sciences
& 38 & 8\,076.03 & 20.68 & 138.02 & 1\,709.82 \\
Digital systems, communication, and media
& 173 & 24\,235.52 & 38.31 & 585.44 & 739.68 \\
Government and public administration
& 203 & 183\,071.43 & 32.61 & 2\,582.07 & 1\,151.45 \\
Industry, infrastructure, and energy
& 109 & 16\,740.21 & 15.46 & 1\,716.80 & 1\,638.44 \\
Mobility and transportation
& 71 & 38\,966.23 & 28.38 & 2\,747.08 & 395.72 \\
\bottomrule
\end{tabular}
\end{subtable}
\medskip
\begin{subtable}{\textwidth}
\centering
\caption{Statistics of the 50 held-out test datasets.}
\label{tab:test-domain-statistics}
\setlength{\tabcolsep}{4pt}
\begin{tabular}{lrrrrr}
\toprule
Domain & \# & \#Cases & \#Activities & Trace length & Duration (days) \\
\midrule
\textbf{Environmental monitoring}
& \textbf{12} & \textbf{64\,923.75} & \textbf{5.42} & \textbf{565.63} & \textbf{2\,170.54} \\
Commerce, finance, and organizations
& 11 & 22\,552.36 & 10.18 & 191.96 & 488.05 \\
Healthcare and life sciences
& 6 & 5\,298.17 & 25.00 & 11.49 & 685.63 \\
Digital systems, communication, and media
& 6 & 162\,867.17 & 10.50 & 4\,266.10 & 166.48 \\
Government and public administration
& 5 & 122\,600.60 & 44.40 & 18.64 & 117.42 \\
Industry, infrastructure, and energy
& 5 & 1\,244.60 & 6.80 & 522.62 & 2\,162.66 \\
Mobility and transportation
& 5 & 408\,464.80 & 42.00 & 3\,709.90 & 50.13 \\
\bottomrule
\end{tabular}
\end{subtable}

\end{table*}

\section{Dataset Information}
\label{sect:dataset-info}

The datasets used for pretraining and testing are collected from a variety of public online repositories, such as
\href{https://data.4tu.nl/}{4TU.ResearchData},
\href{https://www.kaggle.com/datasets}{Kaggle},
\href{https://zenodo.org/}{Zenodo},
\href{https://data.cityofnewyork.us/}{NYC Open Data}, and
\href{https://data.gov/}{Data.gov}.
We process each dataset individually to conform to our event-sequence representation. If a dataset contains more than 100 activity types, we retain the 99 most frequent types and merge the remaining ones into a general ``other'' activity. The test datasets are verified to ensure that they do not overlap with the pretraining data in dataset or process identity, including duplicate or derived versions of the same underlying event log.
The training split of each test dataset---used both for training local models and as the context set for foundation-model inference---is capped at 10,000 traces. If the split contains more than 10,000 traces, we randomly subsample 10,000.

\begin{figure}
    \centering
    \includegraphics[width=0.7\linewidth]{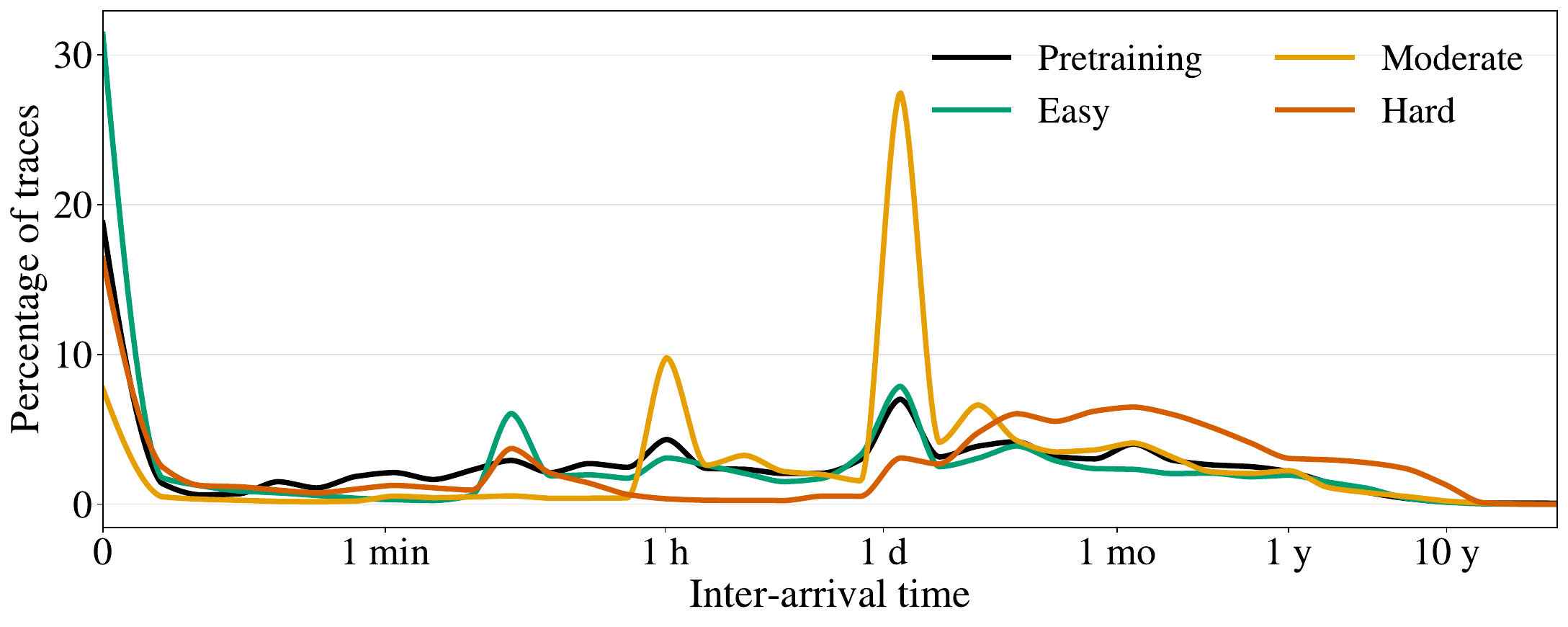}
    \medskip
    \includegraphics[width=0.7\linewidth]{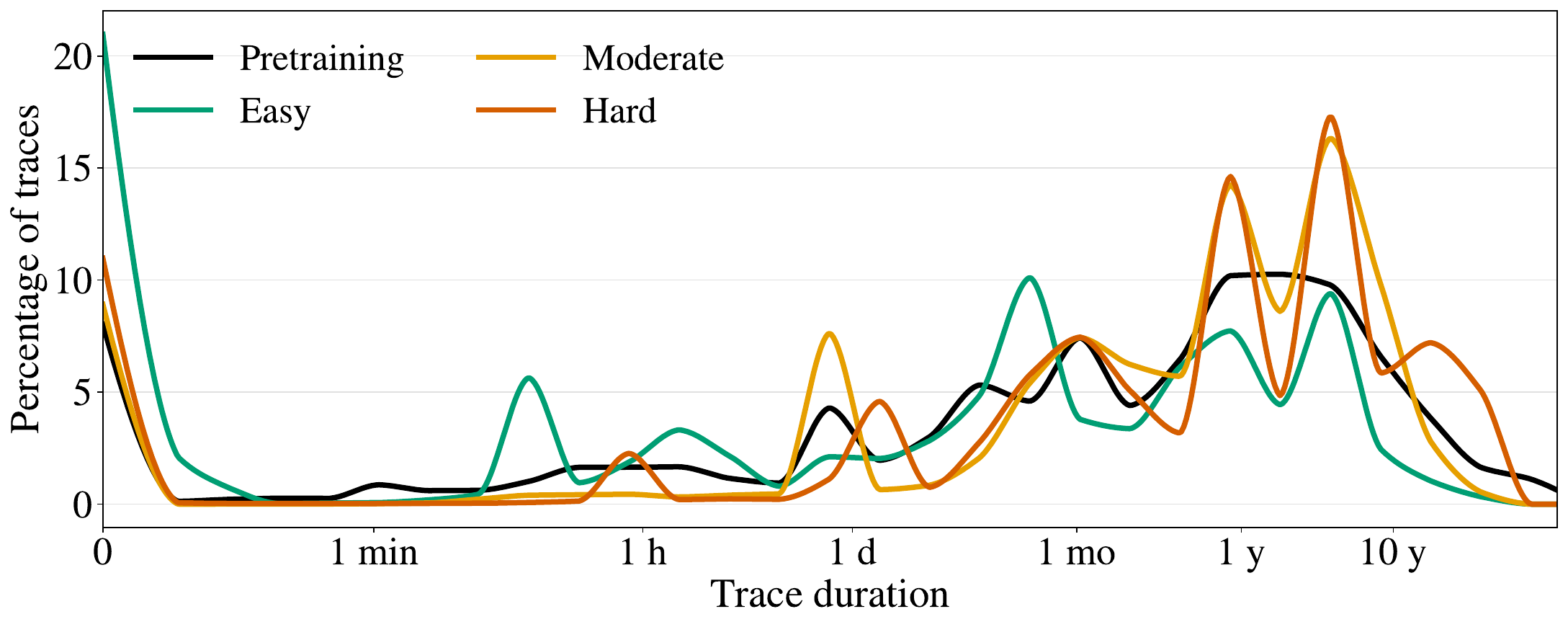}
    \medskip
    \includegraphics[width=0.7\linewidth]{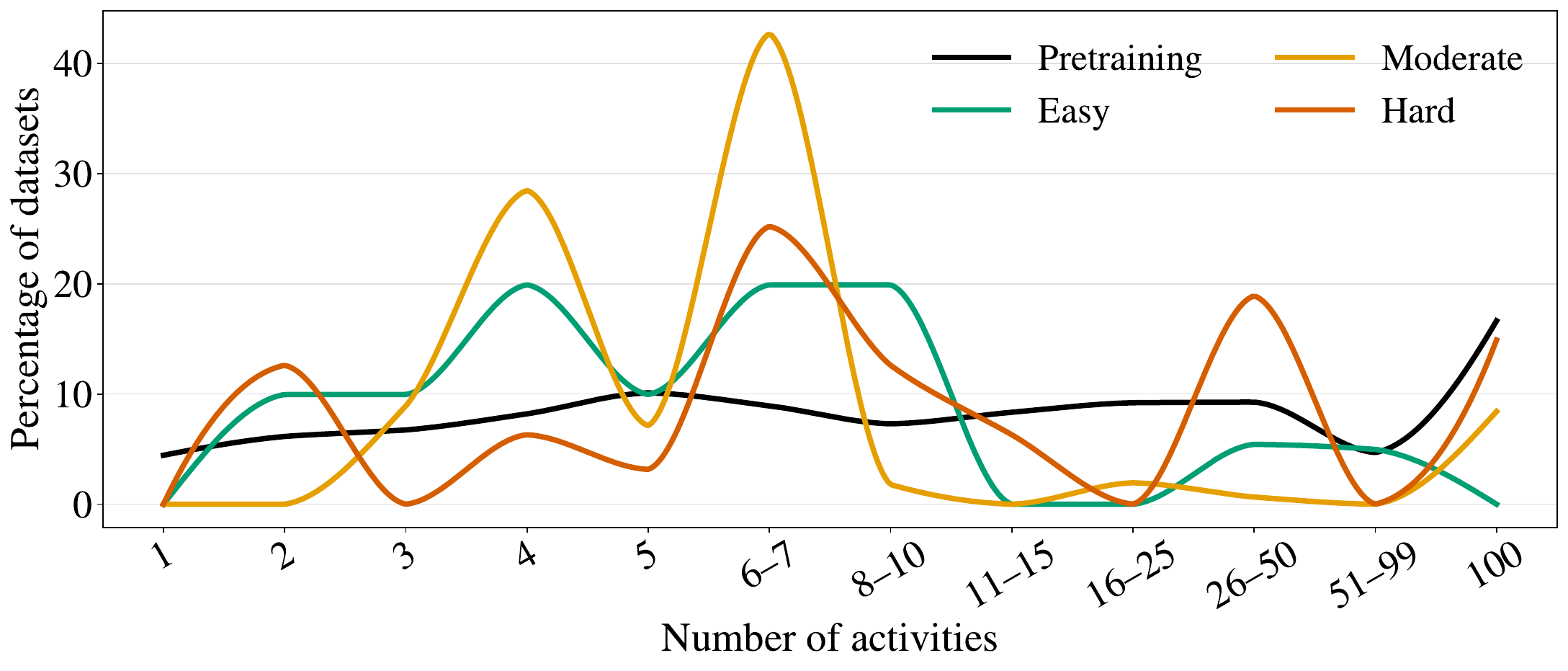}
    \medskip
    \includegraphics[width=0.7\linewidth]{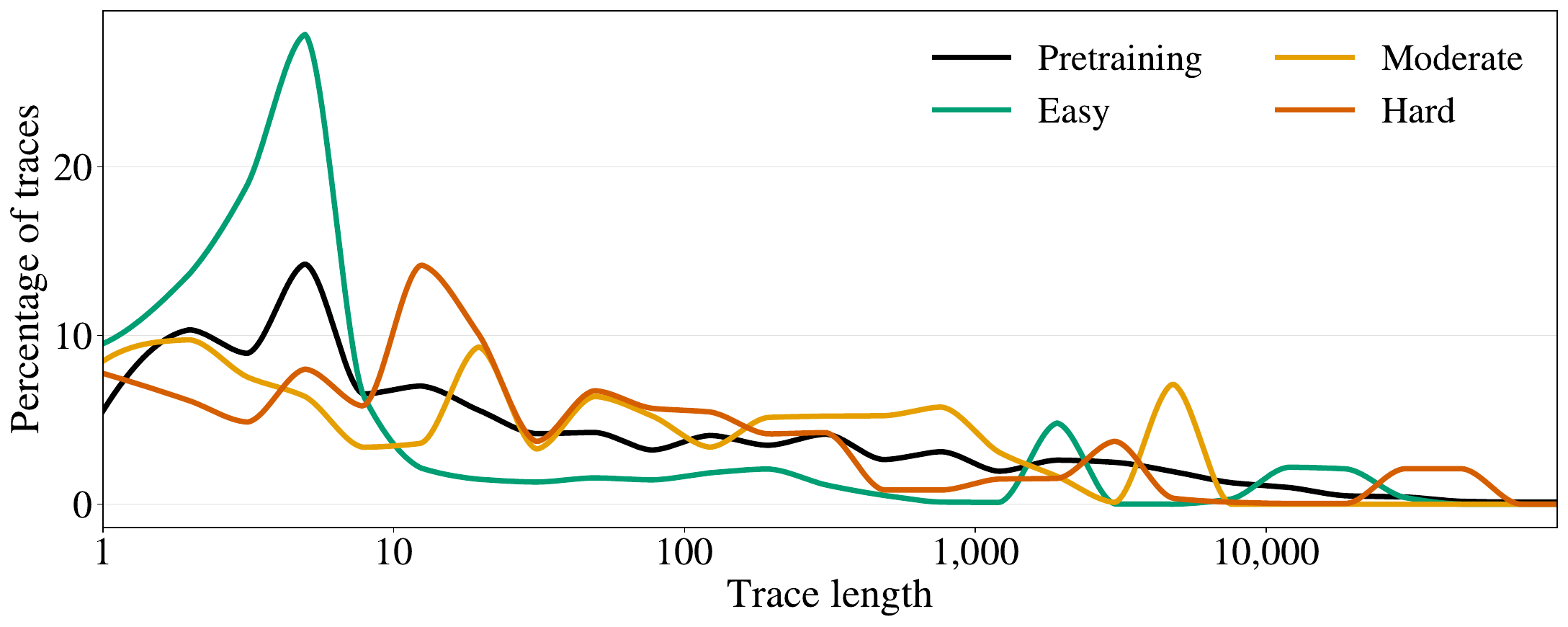}
    \caption{Interarrival times, trace durations, activity counts, and trace lengths per difficulty
    in the test datasets, and among the pretraining datasets. Interarrival times, trace durations,
    and trace lengths are per-dataset averages.}

    \label{fig:train-test-agg-metrics}
\end{figure}

Table~\ref{tab:dataset-domain-statistics} summarizes the pretraining and test datasets by domain. Notably, environmental monitoring is completely absent from the pretraining pool, enabling evaluation on a previously unseen domain. Figure~\ref{fig:train-test-agg-metrics} further compares the distributions of inter-arrival time, trace duration, number of activities, and trace length between the pretraining pool and the easy, moderate, and hard test-dataset groups. The test datasets exhibit {noticeable} distribution shifts relative to the pretraining data, particularly in inter-arrival times and activity counts for the hard group. Together, these differences make the test collection a challenging benchmark for evaluating cross-system generalization of foundation models.

Finally, Table~\ref{tab:test-only-dataset-catalog} provides the full catalog of the 50 test datasets, including a brief description of each dataset and preprocessing details where applicable.

\begingroup
\scriptsize
\begin{longtable}{p{8.5cm}rrrr}
\caption{Details of the 50 real-world test datasets. Trace length and duration (in days) are averaged over all traces within each dataset.}\label{tab:test-only-dataset-catalog}\\
\toprule
Dataset and source & \#Cases & \#Act. & Length & Duration \\
\midrule
\endfirsthead
\toprule
Dataset and source & \#Cases & \#Act. & Length & Duration \\
\midrule
\endhead
\midrule
\multicolumn{5}{r}{Continued on next page} \\
\endfoot
\bottomrule
\endlastfoot
\multicolumn{5}{l}{\emph{Environmental monitoring}} \\[0.1cm]

\textbf{Beijing Air Quality} (\citeauthor{chen2017beijing}, \citeyear{chen2017beijing}) & 12 & 5 & 4\,387 & 1\,218 \\*
\multicolumn{5}{p{13cm}}{Hourly PM2.5 observations from 12 Beijing stations. Values are discretized into six AQI categories and only transitions between categories are retained.} \\[0.5cm]
\textbf{EPA AQS Ozone} (\href{https://aqs.epa.gov/aqsweb/airdata}{EPA AQS}) & 1\,225 & 6 & 186.1 & 205.6 \\*
\multicolumn{5}{p{13cm}}{Hourly ozone measurements from US EPA monitoring sites. Concentrations are mapped to six AQI categories and only changes in category are emitted as events.} \\[0.5cm]
\textbf{EPA AQS PM2.5} (\href{https://aqs.epa.gov/aqsweb/airdata}{EPA AQS}) & 1\,029 & 6 & 57.60 & 290.9 \\*
\multicolumn{5}{p{13cm}}{PM2.5 observations from US EPA monitoring sites. Measurements are mapped to six AQI categories and ordered per monitoring station.} \\[0.5cm]
\textbf{EPA ECHO Enforcement} (\href{https://echo.epa.gov/tools/data-downloads}{EPA ECHO}) & 107\,695 & 6 & 14.90 & 4\,248 \\*
\multicolumn{5}{p{13cm}}{EPA ECHO facility histories covering inspections, detected violations, enforcement actions, penalties, and return to compliance. Source tables are joined into one lifecycle per facility.} \\[0.5cm]
\textbf{EPA ICIS-Air} (\href{https://echo.epa.gov/tools/data-downloads/icis-air-download-summary}{EPA ECHO}) & 138\,133 & 4 & 11.20 & 2\,889 \\*
\multicolumn{5}{p{13cm}}{Clean Air Act compliance histories from EPA ICIS-AIR. Inspections, violation determinations, high-priority violations, and resolutions are combined and ordered per facility.} \\[0.5cm]
\textbf{EPA NPDES} (\href{https://echo.epa.gov/tools/data-downloads/icis-npdes-download-summary}{EPA ECHO}) & 211\,120 & 5 & 9.70 & 2\,055 \\*
\multicolumn{5}{p{13cm}}{NPDES water-compliance histories. Inspection, violation, informal/formal enforcement, and resolution records are combined into chronological facility traces.} \\[0.5cm]
\textbf{EPA RCRA} (\href{https://echo.epa.gov/tools/data-downloads/rcra-download-summary}{EPA ECHO}) & 266\,298 & 3 & 6.60 & 1\,610 \\*
\multicolumn{5}{p{13cm}}{RCRA hazardous-waste compliance histories. Facility evaluations, violation determinations, and enforcement actions are merged and ordered by date.} \\[0.5cm]
\textbf{India Air Quality} (\href{https://www.kaggle.com/datasets/rohanrao/air-quality-data-in-india}{Kaggle}) & 52\,226 & 6 & 22.40 & 0.912 \\*
\multicolumn{5}{p{13cm}}{Hourly Indian air-quality measurements grouped by station and day. PM2.5 is discretized into six AQI categories; each station-day forms one case.} \\[0.5cm]
\textbf{NOAA Coral Reef Watch} (\href{https://coralreefwatch.noaa.gov/satellite/docs/recommendations_crw_citation.php}{NOAA CRW}) & 1\,080 & 10 & 14.10 & 355.6 \\*
\multicolumn{5}{p{13cm}}{NOAA Coral Reef Watch thermal-stress histories. Product observations are represented as ten bleaching or thermal-stress states and ordered for each monitored region-period case.} \\[0.5cm]
\textbf{US Drought Monitor} (\citeauthor{svoboda2002drought}, \citeyear{svoboda2002drought}) & 51 & 6 & 113.3 & 7\,602 \\*
\multicolumn{5}{p{13cm}}{Weekly drought conditions for US states. Area percentages are converted to the most severe drought category exceeding a coverage threshold, and only category changes are retained.} \\[0.5cm]
\textbf{USGS Streamflow} (\href{https://waterservices.usgs.gov/}{USGS NWIS}) & 16 & 4 & 433.1 & 4\,039 \\*
\multicolumn{5}{p{13cm}}{Daily discharge histories for 16 major river gauges. Per-station quantiles define low, normal, high, and flood regimes, and only regime transitions are retained.} \\[0.5cm]
\textbf{USGS Water} (\href{https://waterdata.usgs.gov/nwis}{USGS Water Data}) & 200 & 4 & 1\,532 & 1\,534 \\*
\multicolumn{5}{p{13cm}}{Daily USGS water-discharge histories for 200 gauges. Measurements are mapped to four flow regimes using station-specific thresholds and ordered per gauge.} \\[0.5cm]
\midrule
\multicolumn{5}{l}{\emph{Commerce, finance, and organizations}} \\[0.1cm]
\textbf{Amazon Reviews} (\citeauthor{mcauley2013amateurs}, \citeyear{mcauley2013amateurs}) & 21\,504 & 5 & 9.60 & 371.1 \\*
\multicolumn{5}{p{13cm}}{Amazon food-review histories, with one case per reviewer and rating values as activities. Reviews are ordered by their original timestamps; simultaneous reviews are retained.} \\[0.5cm]
\textbf{Berka Banking} (\citeauthor{berka1999financial}, \citeyear{berka1999financial}) & 4\,500 & 5 & 212.5 & 1\,118 \\*
\multicolumn{5}{p{13cm}}{Bank-account transaction histories; the benchmark exposes operation, transaction type, and purpose prediction tasks, while the displayed statistics use the operation task.} \\[0.5cm]
\textbf{DeFi Protocol Fees} (\href{https://defillama.com/}{DefiLlama}) & 2\,100 & 7 & 444.7 & 498.2 \\*
\multicolumn{5}{p{13cm}}{Daily fee histories for decentralized-finance protocols. Percentage changes are discretized into seven categories ranging from fee collapse to fee surge.} \\[0.5cm]
\textbf{DeFi Protocol TVL} (\href{https://defillama.com/}{DefiLlama}) & 1\,108 & 7 & 795.2 & 804.7 \\*
\multicolumn{5}{p{13cm}}{Daily total-value-locked histories for DeFi protocols. Relative TVL changes are binned into seven states from crash through surge.} \\[0.5cm]
\textbf{DEX Trading Volume} (\href{https://defillama.com/}{DefiLlama}) & 1\,050 & 7 & 606.3 & 658.1 \\*
\multicolumn{5}{p{13cm}}{Daily trading-volume histories for decentralized exchanges. Relative changes are discretized into seven activity levels from volume crash through volume surge.} \\[0.5cm]
\textbf{Employee HR} (\href{https://www.kaggle.com/datasets/ravindrasinghrana/employeedataset}{Kaggle}) & 2\,771 & 2 & 1.40 & 193.6 \\*
\multicolumn{5}{p{13cm}}{Employee lifecycles containing hire and, where observed, exit events. Raw employment dates are converted into chronologically ordered two-state traces per employee.} \\[0.5cm]
\textbf{Hotel Booking Demand} (\citeauthor{antonio2019hotel}, \citeyear{antonio2019hotel}) & 92\,930 & 4 & 2.40 & 68.80 \\*
\multicolumn{5}{p{13cm}}{Hotel reservation lifecycles. Each booking is converted into reservation, cancellation or check-in, and check-out events using the supplied booking and arrival dates.} \\[0.5cm]
\textbf{LeadForge Lead Scoring} (\href{https://www.kaggle.com/datasets/derelictpanda/leadforge-lead-scoring-intro-v1}{Kaggle}) & 4\,560 & 4 & 4.30 & 12.90 \\*
\multicolumn{5}{p{13cm}}{CRM engagement histories for sales leads. Calls, emails, meetings, and demonstrations are retained as four ordered activity types per lead.} \\[0.5cm]
\textbf{NYC Inspections} (\href{https://data.cityofnewyork.us/resource/43nn-pn8j}{NYC Open Data}) & 25\,379 & 37 & 9.80 & 554.2 \\*
\multicolumn{5}{p{13cm}}{NYC restaurant inspection histories, with one case per establishment. Inspection and reinspection descriptions are normalized into 37 ordered activity types.} \\[0.5cm]
\textbf{Olist Brazilian E-Commerce} (\href{https://www.kaggle.com/datasets/olistbr/brazilian-ecommerce}{Kaggle}) & 88\,824 & 4 & 3.90 & 12.60 \\*
\multicolumn{5}{p{13cm}}{Brazilian marketplace order lifecycles. Relational order, approval, shipping, and delivery timestamps are joined and converted into four ordered events per order.} \\[0.5cm]
\textbf{Open Food Facts} (\href{https://world.openfoodfacts.org/}{Open Food Facts}) & 3\,350 & 30 & 21.50 & 1\,077 \\*
\multicolumn{5}{p{13cm}}{Open Food Facts product-addition histories grouped by brand. Raw product categories are consolidated into 30 food-category activities and ordered by creation time.} \\[0.5cm]
\midrule
\multicolumn{5}{l}{\emph{Healthcare and life sciences}} \\[0.1cm]
\textbf{ClinicalTrials.gov} (\href{https://clinicaltrials.gov/}{ClinicalTrials.gov}) & 9\,425 & 8 & 6.80 & 1\,898 \\*
\multicolumn{5}{p{13cm}}{Clinical-trial record lifecycles from registration through study completion and result submission. Available milestone dates are converted to eight ordered event types per trial.} \\[0.5cm]
\textbf{FDA Drug Enforcement} (\href{https://open.fda.gov/apis/drug/enforcement/}{openFDA}) & 842 & 3 & 17.30 & 536.8 \\*
\multicolumn{5}{p{13cm}}{FDA drug-recall histories grouped by recalling firm. Recall classifications are represented as Class I, II, or III activities and ordered by report date.} \\[0.5cm]
\textbf{FDA FAERS} (\href{https://open.fda.gov/apis/drug/event/}{openFDA}) & 2\,042 & 50 & 19.80 & 552.0 \\*
\multicolumn{5}{p{13cm}}{FDA adverse-event reports grouped by suspect drug. Reactions are restricted to the 50 most frequent MedDRA preferred terms and ordered by report date.} \\[0.5cm]
\textbf{FitLife} (\href{https://www.kaggle.com/datasets/jijagallery/fitlife-health-and-fitness-tracking-dataset}{Kaggle}) & 12\,000 & 10 & 18.40 & 1\,110 \\*
\multicolumn{5}{p{13cm}}{Participant fitness histories containing ten workout types. Sessions are ordered by their recorded dates;} \\[0.5cm]
\textbf{Medical Appointments} (\href{https://www.kaggle.com/datasets/carogonzalezgaltier/medical-appointment-scheduling-system}{Kaggle}) & 6\,830 & 4 & 2.90 & 16.80 \\*
\multicolumn{5}{p{13cm}}{Patient appointment lifecycles. Scheduling records are converted into scheduled, check-in, completed, and no-show events and ordered per patient.} \\[0.5cm]
\textbf{MIMIC-II} (\citeauthor{saeed2011mimic2}, \citeyear{saeed2011mimic2}) & 650 & 75 & 3.72 & 0.0000201 \\*
\multicolumn{5}{p{13cm}}{Temporal-point-process benchmark with 75 diagnosis/event types. We retain its published preprocessing and scaled event times rather than rebuilding traces from raw MIMIC tables.} \\[0.5cm]
\midrule
\multicolumn{5}{l}{\emph{Digital systems, communication, and media}} \\[0.1cm]
\textbf{ATP Tennis Tournaments} (\href{https://github.com/JeffSackmann/tennis_atp}{Tennis Abstract}) & 811\,988 & 32 & 1.90 & 0 \\*
\multicolumn{5}{p{13cm}}{ATP player--tournament journeys. Match records are converted to round/outcome events (for example, quarter-final won or lost) ordered within each tournament journey.} \\[0.5cm]
\textbf{CIC-IDS2017} (\citeauthor{sharafaldin2018toward}, \citeyear{sharafaldin2018toward}) & 3 & 6 & 25\,421 & 1.40 \\*
\multicolumn{5}{p{13cm}}{Network-security traffic from CIC-IDS2017. Grouped into host-level temporal traces and six attack/activity classes; the resulting traces are segmented into training examples.} \\[0.5cm]
\textbf{Enron Email} (\href{https://enrondata.org/en/latest/data/calo-enron-email-dataset/}{CALO Enron}) & 1\,262 & 6 & 14.40 & 128.1 \\*
\multicolumn{5}{p{13cm}}{A case-centric view of the Enron email corpus, with one case per email address. Message actions are mapped to six email activities such as default, response, forwarding, and reminder.} \\[0.5cm]
\textbf{MovieLens 25M} (\citeauthor{harper2015movielens}, \citeyear{harper2015movielens}) & 159\,025 & 10 & 152.5 & 216.8 \\*
\multicolumn{5}{p{13cm}}{MovieLens user rating histories. Each user is a case and half-star ratings from 0.5 to 5 are treated as ten categorical activities ordered by rating timestamp.} \\[0.5cm]
\textbf{Stack Exchange Questions} (\href{https://stackoverflow.com/help/licensing}{Stack Exchange}) & 933 & 7 & 5.00 & 651.5 \\*
\multicolumn{5}{p{13cm}}{Stack Exchange question lifecycles. Question, answer, comment, acceptance, closure, and related actions are combined into seven ordered event types per question.} \\[0.5cm]
\textbf{TensorFlow GitHub} (\href{https://github.com/tensorflow/tensorflow}{GitHub}) & 3\,992 & 2 & 1.50 & 1.10 \\*
\multicolumn{5}{p{13cm}}{TensorFlow GitHub issue lifecycles. API records are converted into opened and closed events per issue and ordered by their GitHub timestamps.} \\[0.5cm]
\midrule
\multicolumn{5}{l}{\emph{Government and public administration}} \\[0.1cm]
\textbf{BPI Challenge 2015} (\citeauthor{vandongen2015bpi}, \citeyear{vandongen2015bpi}) & 1\,097 & 100 & 43.00 & 94.70 \\*
\multicolumn{5}{p{13cm}}{Environmental-permit workflows from five Dutch municipalities. We retain the original event ordering and use full activity codes.} \\[0.5cm]
\textbf{US Congressional Bills} (\href{https://www.govinfo.gov/bulkdata/BILLSTATUS}{GovInfo}) & 63\,928 & 9 & 5.20 & 44.00 \\*
\multicolumn{5}{p{13cm}}{US congressional bill histories. Detailed BILLSTATUS actions are mapped to nine legislative stages, such as introduction, committee, floor, presidential action, and enactment.} \\[0.5cm]
\textbf{NSF Awards} (\href{https://www.nsf.gov/awardsearch/download}{NSF}) & 47\,140 & 10 & 3.20 & 345.7 \\*
\multicolumn{5}{p{13cm}}{National Science Foundation award histories grouped by principal investigator. Funding programs are used as the activity, restricted to ten aggregated program groups in this benchmark.} \\[0.5cm]
\textbf{NYC 311} (\href{https://data.cityofnewyork.us/resource/erm2-nwe9}{NYC Open Data}) & 500\,000 & 3 & 2.30 & 17.90 \\*
\multicolumn{5}{p{13cm}}{NYC 311 service-request lifecycles. Requests are sampled to 500,000 cases and represented by created, in-progress, and closed events inferred from available status timestamps.} \\[0.5cm]
\textbf{WABO Permits} (\href{https://data.4tu.nl/datasets/a928c371-d8e8-4c14-95db-2c28078085b5}{4TU.ResearchData}) & 838 & 100 & 39.50 & 84.80 \\*
\multicolumn{5}{p{13cm}}{Receipt-phase environmental-permit workflows from a Dutch municipality. Original detailed process codes and timestamps are retained and ordered per permit application.} \\[0.5cm]
\midrule
\multicolumn{5}{l}{\emph{Industry, infrastructure, and energy}} \\[0.1cm]
\textbf{Backblaze SMART} (\href{https://www.backblaze.com/cloud-storage/resources/hard-drive-test-data}{Backblaze}) & 4\,334 & 4 & 2.20 & 34.10 \\*
\multicolumn{5}{p{13cm}}{Daily hard-drive SMART observations, with one case per drive. We discretize into healthy, minor degradation, major degradation, and failure, and repeated unchanged states are collapsed.} \\[0.5cm]
\textbf{EIA Electricity Generation} (\href{https://www.eia.gov/electricity/data/state/}{EIA}) & 308 & 13 & 143.0 & 9\,904 \\*
\multicolumn{5}{p{13cm}}{Annual US electricity-generation records grouped by state and producer type. Fuel or energy source is used as the event activity and records are ordered chronologically.} \\[0.5cm]
\textbf{NASA SMAP/MSL} (\citeauthor{hundman2018spacecraft}, \citeyear{hundman2018spacecraft}) & 81 & 6 & 2\,452 & 21.50 \\*
\multicolumn{5}{p{13cm}}{NASA SMAP and Mars Science Laboratory telemetry channels. Continuous telemetry is converted into six nominal/deviation states, producing one ordered trace per channel.} \\[0.5cm]
\textbf{Pipeline Incidents} (\href{https://www.phmsa.dot.gov/data-and-statistics/pipeline/distribution-transmission-gathering-lng-and-liquid-accident-and-incident-data}{PHMSA}) & 192 & 7 & 12.40 & 854.2 \\*
\multicolumn{5}{p{13cm}}{PHMSA hazardous-liquid pipeline incidents grouped by operator. Detailed cause fields are consolidated into seven incident-cause activities and ordered by incident date.} \\[0.5cm]
\textbf{UCI SECOM} (\citeauthor{mccann2008secom}, \citeyear{mccann2008secom}) & 1\,308 & 4 & 3.00 & 0.00694 \\*
\multicolumn{5}{p{13cm}}{Semiconductor manufacturing cases. Each wafer is represented by inspection start, measurement completion, and pass/fail classification events separated by fixed five-minute offsets.} \\
\midrule
\multicolumn{5}{l}{\emph{Mobility and transportation}} \\[0.1cm]
\textbf{BART Ridership} (\href{https://www.bart.gov/about/reports/ridership}{BART}) & 50 & 2 & 15\,966 & 15.60 \\*
\multicolumn{5}{p{13cm}}{Hourly BART passenger flows, with one case per station. Origin--destination counts are represented as arrival and departure events at their corresponding station and hour.} \\[0.5cm]
\textbf{BTS On-Time Flights} (\href{https://transtats.bts.gov}{BTS}) & 2\,019\,872 & 6 & 4.00 & 0.138 \\*
\multicolumn{5}{p{13cm}}{Domestic US flight lifecycles. Scheduled records are converted into gate departure, takeoff, landing, gate arrival, cancellation, and diversion events for each flight.} \\[0.5cm]
\textbf{GeoLife GPS} (\href{https://www.microsoft.com/en-us/research/project/geolife-building-social-networks-using-human-location-history/}{Microsoft GeoLife}) & 142 & 100 & 1\,004 & 145.1 \\*
\multicolumn{5}{p{13cm}}{Microsoft GeoLife user trajectories. GPS points are mapped to roughly 1-km grid cells, consecutive repeats are collapsed, and all but the 100 most visited cells are mapped to OTHER.} \\[0.5cm]
\textbf{NYC Taxi} (\href{https://www.nyc.gov/site/tlc/about/tlc-trip-record-data.page}{NYC TLC}) & 255 & 2 & 1\,561 & 75.60 \\*
\multicolumn{5}{p{13cm}}{NYC yellow-taxi zone activity. A sample of trips is converted into pickup and drop-off events and regrouped as chronological traces for each taxi zone.} \\[0.5cm]
\textbf{OpenSky Flights} (\href{https://zenodo.org/records/7323875}{Zenodo}) & 22\,005 & 100 & 14.10 & 14.20 \\*
\multicolumn{5}{p{13cm}}{OpenSky aircraft route histories for October 2022. Destination airports form the activities; the 100 most frequent are retained and the remainder are mapped to OTHER.} \\[0.5cm]
\end{longtable}
\endgroup

\end{document}